\documentclass{article}

\usepackage{PRIMEarxiv}

\usepackage[utf8]{inputenc} 
\usepackage[T1]{fontenc}    
\usepackage{hyperref}       
\usepackage{url}            
\usepackage{booktabs}       
\usepackage{amsfonts}       
\usepackage{nicefrac}       
\usepackage{microtype}      
\usepackage{lipsum}
\usepackage{fancyhdr}       
\usepackage{graphicx}       
\usepackage{multirow}
\usepackage{colortbl}
\usepackage{hhline}
\usepackage{color, xcolor}
\usepackage{ulem}
\graphicspath{{media/}}     

\title{PDM-SSD: Single-Stage Three-Dimensional Object Detector With Point Dilation
\thanks{\textit{\underline{Citation}}: 
\textbf{Authors. Title. Pages.... DOI:000000/11111.}} 
}

\author{
  Ao Liang, Haiyang Hua, Jian Fang, Wenyu Chen Huaici Zhao* \\
  Key Laboratory of Opto-Electronic Information Processing, Chinese Academy of Sciences, 110016, Shenyang\\
  Shenyang Institute of Automation, Chinese Academy of Sciences, 110016, Shenyang \\
  University of Chinese Academy of Sciences, 100049, Beijing \\
  \texttt{\{liangao, hczhao\}@sia.cn} \\
}

\begin{document}
\maketitle

\begin{abstract}
One of the important reasons why grid/voxel-based three-dimensional (3D) object detectors can achieve robust results for sparse and incomplete targets in Light Detection And Ranging (LiDAR) scenes is that the repeated padding, convolution, and pooling layers in the feature learning process enlarge the model's receptive field, enabling features even in space not covered by point clouds. However, they require time- and memory-consuming 3D backbones. Point-based detectors are more suitable for practical application, but current detectors can only learn from the provided points, with limited receptive fields and insufficient global learning capabilities for such targets. In this paper, we present a novel Point Dilation Mechanism for single-stage 3D detection (PDM-SSD) that takes advantage of these two representations. Specifically, we first use a PointNet-style 3D backbone for efficient feature encoding. Then, a neck with Point Dilation Mechanism (PDM) is used to expand the feature space, which involves two key steps: point dilation and feature filling. The former expands points to a certain size grid centered around the sampled points in Euclidean space. The latter fills the unoccupied grid with feature for backpropagation using spherical harmonic coefficients and Gaussian density function in terms of direction and scale. Next, we associate multiple dilation centers and fuse coefficients to obtain sparse grid features through height compression. Finally, we design a hybrid detection head for joint learning, where on one hand, the scene heatmap is predicted to complement the voting point set for improved detection accuracy, and on the other hand, the target probability of detected boxes are calibrated through feature fusion. On the challenging Karlsruhe Institute of Technology and Toyota Technological Institute (KITTI) dataset, PDM-SSD achieves state-of-the-art results for multi-class detection among single-modal methods with an inference speed of 68 frames. We also demonstrate the advantages of PDM-SSD in detecting sparse and incomplete objects through numerous object-level instances. Additionally, PDM can serve as an auxiliary network to establish a connection between sampling points and object centers, thereby improving the accuracy of the model without sacrificing inference speed. Our code will be available at \url{https://github.com/AlanLiangC/PDM-SSD.git}.\end{abstract}


\section{Introduction}
LiDAR (Light Detection and Ranging) is an active sensor with excellent anti-interference capability, and its output point cloud can provide an accurate 3D representation of the scene. 3D object detection from point clouds has become increasingly popular thanks to its wide applications, such as autonomous driving and virtual reality. Currently, many point cloud-based 3D detection models have been proposed and achieved state-of-the-art performance on various public datasets, such as KITTI \cite{geiger2012we} and Waymo \cite{sun2020scalability}.

The sparsity of point clouds is the main characteristic that distinguishes 3D object detection from traditional 2D detection, and efficiently representing the sparse and unordered point clouds is the key for subsequent processing \cite{fan2021rangedet}. The main approach is to transform the original point clouds into regular feature representations, including projecting the point clouds onto 2D images from a bird's-eye view (BEV) or frontal view (FV), or transforming them into dense 3D voxels \cite{simon_complex-yolo_2018,noauthor_multi-view_nodate,beltran2018birdnet,zeng2018rt3d,ali2018yolo3d,barrera2020birdnet+,CHEN2023110952}. Then, well-performing feature extractors and detectors in 2D image tasks can be directly used for 3D object detection. BEV and FV feature maps can be obtained from voxels and pillars by projecting the 3D feature. The feature maps are always dense 2D representations, where each pixel corresponds to a specific region and encodes the points' information in this region. Voxel-based detection methods preserve the spatial information of the original point clouds to a great extent based on the size of voxels, and using 3D convolutional neural networks can achieve desirable detection performance. Another stream of techniques follows the point-based pipeline to directly operate on raw point clouds \cite{zheng2021se, zhang2022not, shi2019pointrcnn, shi2020point, yang20203dssd, hu2021learning, hu2021sqn, hu2022sensaturban, wei2022spatial}. These detectors construct symmetric functions to deal with the unordered nature of point clouds and expand the receptive field through continuous downsampling and aggregation of local features to obtain rich spatial and semantic features.

\begin{figure}[t]
	\begin{center}
		\includegraphics[width=.5\textwidth]{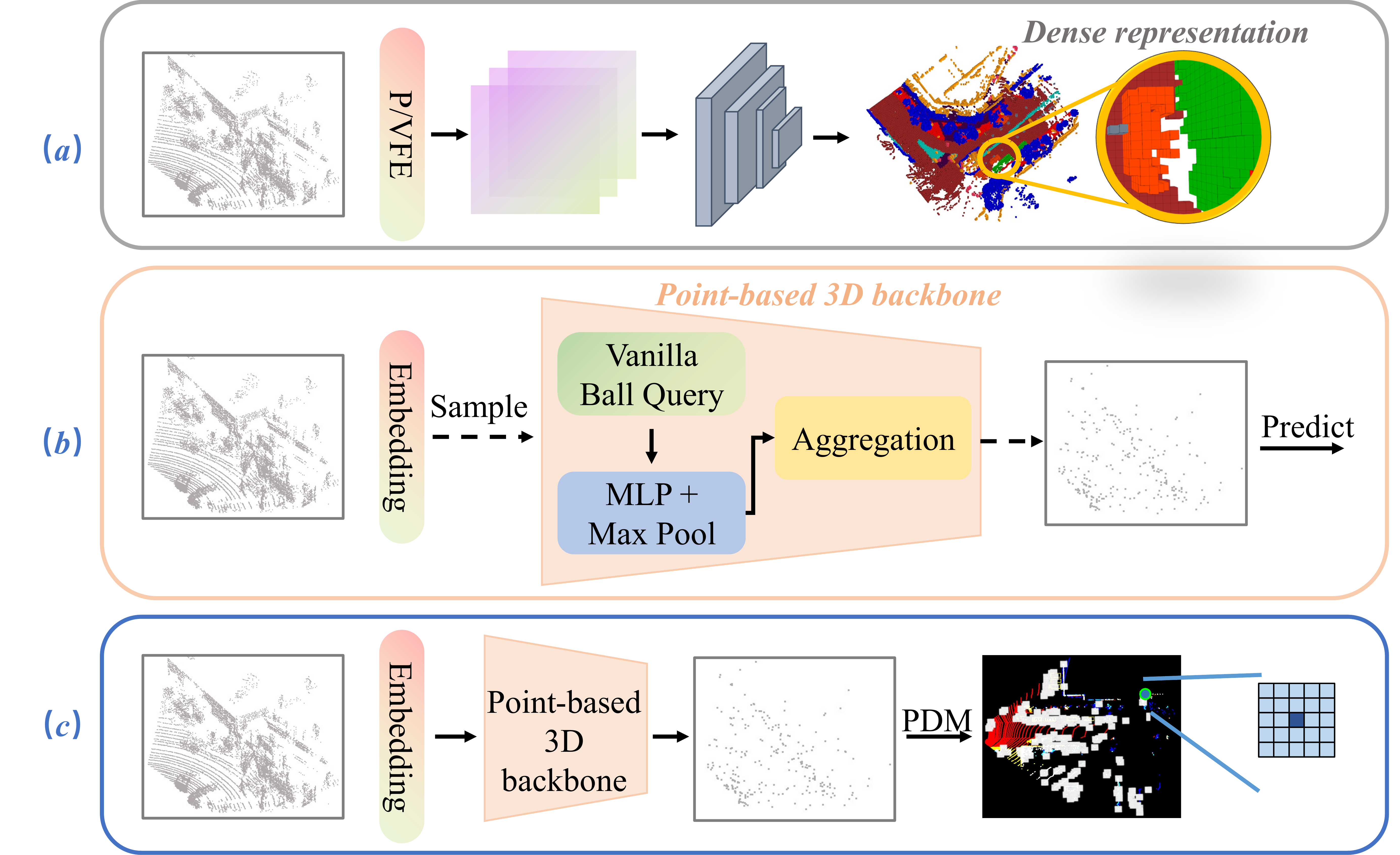}
	\end{center}
	\vspace{-0.3cm}
	\caption{(a) The basic structure of the Grid-based 3D detector. P/VFE means Pillar/Voxel Feature Encoder. This approach allows for obtaining dense feature maps from sparse point clouds as input. (b) The Point-based detector has a basic structure where sparse point clouds are inputted and undergo multiple stages of downsampling, feature learning, and local feature aggregation to obtain sparser point-wise features. (c) The basic structure of our PDM-SSD. Point-wise features obtained from the point-based 3D backbone are lifted to the grid level through PDM. This joint learning approach helps alleviate the limited receptive field problem in (b).}
	\label{fig1}
	\vspace{-0.3cm}
\end{figure}

In general, both grid-based and point-based detectors have distinct advantages and disadvantages. Specifically, grid-based detectors map all points to a regular feature space, which reduces the cost of model design. The padding operation in the convolution process also fills the unused space in the original point cloud with features. Through joint learning with dense detection heads like the center head \cite{yin2021center}, the grid-based detector can capture the overall features of the targets, making it robust to incomplete objects. However, these detectors require complex and parameter-heavy feature encoders, which puts them at a disadvantage in terms of inference speed. Additionally, due to the wide scene range and sparse point cloud in 3D object detection tasks, grid-based feature representations often contain many non-occupied grids, leading to computational waste. Although 2/3D sparse convolution can effectively address these issues, many inference libraries like TensorRT lack operators for sparse models, making their deployment challenging.

The point-based detector does not lose spatial information and theoretically can learn objects of any size and shape. Moreover, this type of detector does not contain operators like sparse convolution in its backbone, making it more suitable for practical deployment in terms of inference speed and memory consumption. However, these models can only extract information from existing points, and for objects with only local regions containing points, their receptive field cannot increase with the increase of the query radius. The feature space is limited to the regions with points, which makes them less robust in detecting sparse objects compared to grid-based detectors. Additionally, the challenge of point cloud sparsity is further amplified in point-based detection models, as the number of object points decreases significantly after progressive downsampling, leading to the loss of object features (especially for small objects) and the potential retention of noise (caused by sensors). This is also the main reason why the detection performance of point-based detectors is lower than that of grid-based detectors at present. Although works like 3D-SSD \cite{yang20203dssd} and IA-SSD \cite{zhang2022not} have improved the recall rate of foreground points by clever downsampling strategies, the issue of discontinuous receptive field in point-based detectors has not received enough attention.

Motivated by this, in this paper, we propose PDM-SSD that combines the advantages of both grid and point representations while overcoming their drawbacks. Instead of projecting points onto a regular grid, we directly use a PointNet-style 3D backbone to extract features from the original point cloud. This type of feature extractor has a small number of parameters and theoretically no spatial information loss. After several rounds of downsampling and feature aggregation, each sampled point stores both geometric and semantic information within its receptive field. Inside the PDM-SSD, sampled points are adaptively lifted onto a 2D grid using our Point Dilation Mechanism. By using spherical harmonics coefficients and Gaussian density functions, we fill the 2D grid with initial features in terms of direction and scale. In this way, the space that is not occupied by the original point cloud is also included in the learning scope of the model. By filling the initial features, subsequent feature learning is also supported. As both the point-wise features without spatial information loss and the grid-features with continuous receptive fields are aggregated, we design a hybrid detection head that can fully utilize these two types of features to decode the features. This allows PDM-SSD to improve the detection capability for incomplete objects while maintaining inference speed.  Extensive experiments have been conducted on the KITTI detection benchmark to verify the effectiveness and efficiency of our approach. PDM-SSD outperforms all state-of-the-art point-based single stage methods for multi-class detection and performs comparably to two-stage point-based and grid-based methods as well. To summarize, the contributions are listed as follows:

\begin{itemize}
	\item {We have evaluated the advantages and disadvantages of existing grid-based and point-based detectors, and proposed the Point Dilation Mechanism to combine the strengths of both representations. To the best of our knowledge, this is the first point-based method that addresses the issue of discontinuous receptive fields.}
	\item {We introduce a novel point-based single stage 3D detector, PDM-SSD, which surpasses all state-of-the-art point-based single stage methods for multi-class detection on the KITTI benchmark.}
	\item {Our proposed PDM-SSD strikes a good balance between accuracy and inference speed, making it a deployment-friendly model. Furthermore, we demonstrate that even when the Point Dilation Mechanism is used as an auxiliary network, significant benefits can still be achieved.}
\end{itemize}

The structure of this paper is as follows. Section \ref{sec:related} provides a review of existing works on LiDAR-based detectors. In Section \ref{sec:method}, we explicitly formulate the changes in receptive field of point-based and grid-based models during the process of feature learning, followed by a technical implementation of PDM-SSD. In Section \ref{sec:experiments}, we conduct experiments on the KITTI dataset to demonstrate the effectiveness of our method, and we perform ablation experiments to validate the rationality of PDM-SSD design. In addition, we also use a large number of object-level instances to illustrate the contributions of PDM-SSD in detecting difficult targets. Finally, Section \ref{sec:conclusion} concludes this paper and provides an outlook on future work.

\section{Related Work}
\label{sec:related}
In this paper, we focus on single model 3D object detectors based on pure point clouds. Based on the structural types of point clouds input into the 3D object detectors, we provide a brief overview.

\subsection{Grid-based 3D object detectors}
Grid-based 3D object detectors first convert point cloud into discrete grid representations such as voxels, pillars, BEV feature maps, and FV feature maps. Voxels are 3D cubes that contain points inside voxel cells. They can preserve the spatial information of the original point cloud to a great extent while reducing the number of computational units. VoxelNet \cite{zhou_voxelnet_2017} is a pioneering work that uses sparse voxel grids and proposes a novel voxel feature encoding (VFE) layer to extract features from the points inside a voxel cell. Building upon VoxelNet, VoxelNeXt \cite{chen2023voxelnext} utilizes a fully sparse structure for feature extraction and detection heads, partially overcoming the high memory consumption issue of VoxelNet. Currently, most state-of-the-art detectors on public datasets use voxel-based feature extractors, such as DistillBEV \cite{wang2023distillbev} and LidarMultiNet \cite{ye_lidarmultinet_2022}. Pillars can be seen as special voxels with a voxel size of 1 in the vertical direction. PointPillars \cite{lang_pointpillars_2019} is a seminal work that introduces the pillar representation, and then PillarNet \cite{shi_pillarnet_2022} and FastPillars \cite{zhou_fastpillars_2023} add additional encoder networks to enhance feature learning in the model structure. Compared to voxel-based detectors, pillar-based ones have fewer grid cells, making them more similar to 2D image detection and enabling faster inference speed.
The Bird’s-eye view (BEV) feature map is a dense 2D representation typically obtained by projecting voxel, pillar, and raw point cloud data. BEV-based detectors \cite{simon_complex-yolo_2018,noauthor_multi-view_nodate,beltran2018birdnet,zeng2018rt3d,ali2018yolo3d,barrera2020birdnet+} usually require the addition of cell-level features to expand the feature space, such as height and density. The feature learning process for those detectors is similar to 2D detection tasks. It is worth noting that the BEV representation of the scene and perception results greatly simplifies the environmental space in which the objects are located, making it easy to propagate information to downstream tasks such as path planning and control \cite{jia2023driveadapter}. Therefore, BEV-based detectors are currently a hot research direction and are gradually forming a new generation of BEV perception paradigm with multiple tasks and modalities \cite{hu2023planning,li2023delving,li2022bevformer}.
The range image is a dense 2D representation that contains 3D distance information in each pixel. LaserNet \cite{meyer2019lasernet} is a pioneering work that detects 3D objects from range images, utilizing the deep layer aggregation network (DLA-Net) \cite{yu2018deep} to extract multi-scale features. Laserflow \cite{meyer2020laserflow} and RangeRCNN \cite{liang2020rangercnn} further enhance feature learning by adding an encoder-decoder structure. However, when preprocessing point clouds into Bird's Eye View (BEV) and Front View (FV) feature maps, spatial information is lost. This makes it more challenging to identify objects with pillar-like features, such as pedestrians, in the BEV representation. Additionally, in the RV representation, occlusions make it harder to accurately localize objects. In general, grid-based detectors have prominent advantages and disadvantages. The regularized feature representation significantly reduces the design cost of the model, and models that perform well in 2D detection tasks can be easily transferred to 3D tasks. Furthermore, through padding, convolution, and pooling operations, the unoccupied space in the original point cloud is filled with features, making the model more robust in object recognition. However, such models compress the geometric information of the point cloud while regularizing it. To extract deep features, a backbone with a large number of parameters is required, which greatly increases the deployment cost and reduces the inference speed of the model. Moreover, since point clouds are sparsely distributed, most grid cells in 3D space are empty and do not contain any points, leading to high memory consumption in grid-based methods, further increasing the deployment cost of the model.

\begin{figure}[t]
	\begin{center}
		\includegraphics[width=.8\textwidth]{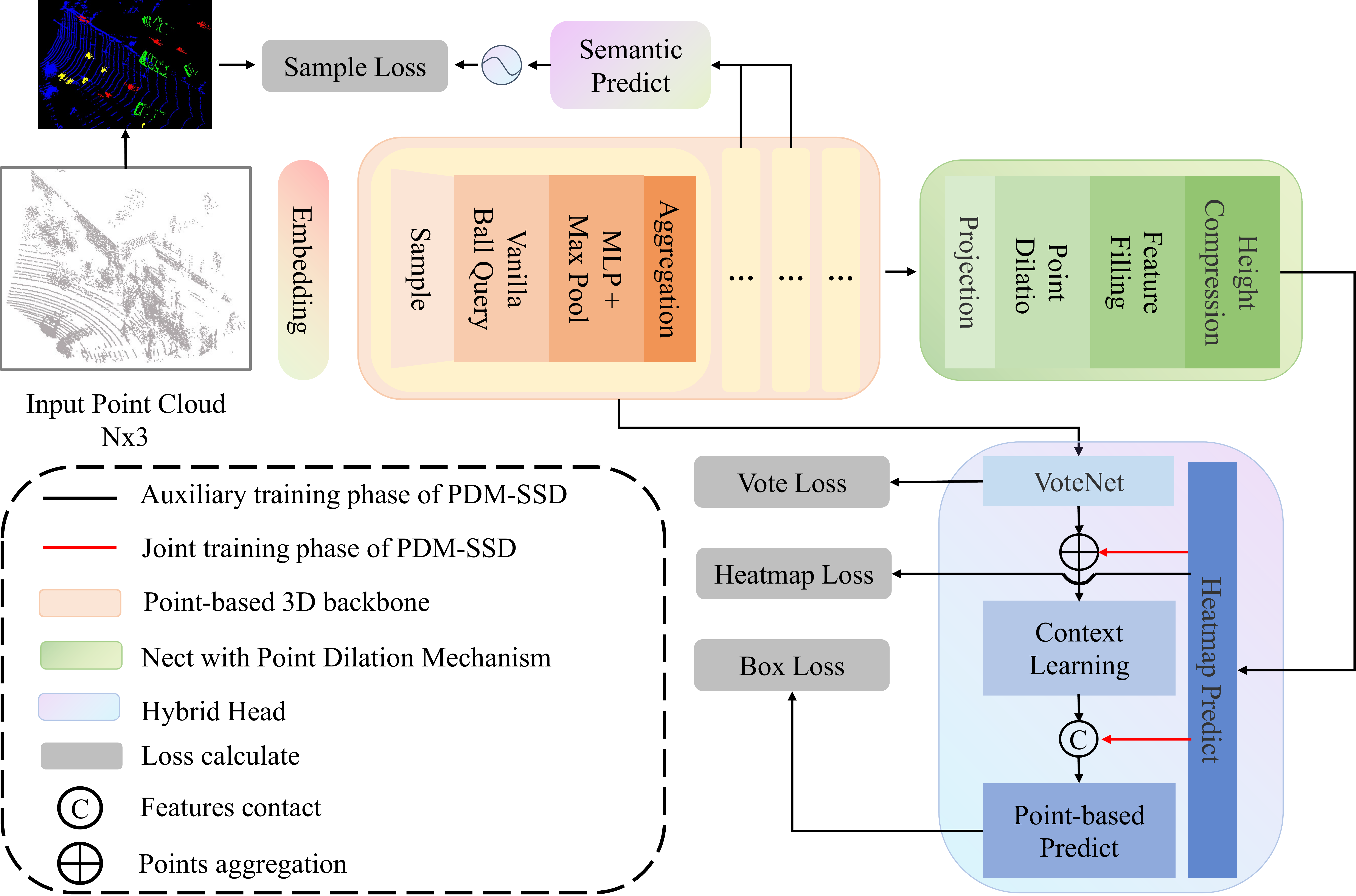}
	\end{center}
	\vspace{-0.3cm}
	\caption{The overall workflow of PDM-SSD. In the joint training phase, the input LiDAR point clouds are first passed through the embedding network to expand the feature space of the points. Then, a PointNet-style 3D backbone network is utilized to extract features for each point. This 3D backbone network consists of several stages of downsampling modules, local feature aggregation modules, and multi-scale feature aggregation modules. The neck network includes our proposed point dilation mechanism, where points are lifted to the grid level and feature filling is performed for the unoccupied space in the original point cloud using a special mechanism. The grid-wise features are then used to regress the heatmap of the scene, which provides information about the target's position, and the grid features are jointly learned with the fusion detection head to learn the global features of the target. In the auxiliary training phase, we do not utilize the information provided by neck but only compute the prediction loss of heatmap.}
	\label{fig2}
	\vspace{-0.3cm}
\end{figure}

\subsection{Point-based 3D object detectors}
Different from grid-based methods, point-based methods \cite{qi_deep_2019,shi2019pointrcnn,yang_3dssd_2020} directly learn geometry from unstructured point clouds. Specifically, point clouds are first passed through a point-based backbone network, where points are gradually sampled and features are learned by point cloud operators, further generating specific proposals for objects of interest. 3D bounding boxes are then predicted based on the features of these proposals. The implementation of point-based detectors relies on the early work on point-based point cloud classifiers. PointNet \cite{qi2017pointnet} pioneers this class of methods, which consists of two basic structures: a multilayer perceptron (MLP) network for feature learning and max-pooling as a symmetric function. The former can transform features to specified dimensions and learn global features of the point cloud, while the latter can overcome the disorder of point clouds, and its symmetry ensures that the model has the same output regardless of the input order.
Currently, the main differences among point-based detectors lie in the downsampling method and the structure of the local feature extractor. To ensure global coverage of the sampled points, PointRCNN \cite{shi2019pointrcnn} pioneers the use of Furthest Point Sampling (FPS) to progressively downsample the input point cloud and generate 3D proposals. To improve the recall rate of foreground points and reduce information loss of the objects, 3DSSD \cite{yang_3dssd_2020} proposed the Feature Furthest Point Sampling (F-FPS) method, which uses feature distance instead of Euclidean distance to adaptively retain foreground points. Zhang et al. \cite{zhang2022not} proposed an instance-aware sampling method, which adds a branch network to predict the semantic information of points in the sampling process to retain foreground points. Their lightweight object detector, IA-SSD, greatly improves the recall rate of foreground points and achieves state-of-the-art accuracy on the KITTI dataset. However, IA-SSD relies heavily on dataset-specific hyperparameters, and this hard-labeling method leads to feature redundancy. Yang et al. \cite{yang2022dbq} proposed DBQ-SSD to dynamically aggregate local features, further improving the inference speed without sacrificing detection accuracy. Liang et al. \cite{liang2023spsnet} believed that foreground points contribute differently to the detection results and proposed SPSNet to regress the stability of the sampled points, thereby sequentially introducing a stability-based downsampling method. 
In summary, point-based detectors directly take raw point clouds as input without loss of spatial information. The feature learning module has a small number of parameters, making it more suitable for practical applications in terms of inference speed and memory consumption. For example, IA-SSD and DBQ-SSD achieve detection speeds of 83FPS and 162FPS, respectively, on an 2080Ti GPU. However, due to the unordered and irregular nature of point clouds, the difficulty of model design is much greater than that of convolutional and transformer-based models. Additionally, point-based detectors can only extract features from the provided point cloud space. For incomplete and sparse objects, their perception range does not change as the query ball radius increases, resulting in discontinuous receptive fields. It is difficult for them to learn the overall features of the objects. Therefore, point-based detectors are far less robust than grid-based detectors \cite{zhu2023understanding} in detecting diluted objects caused by occlusion and adverse weather conditions.

\subsection{Point-voxel based 3D object detectors}
Point-voxel based approaches combine points and voxels in the 3D object detection process to overcome the limitations of point-based and voxel-based methods. Liu \textit{et al.} \cite{liu2019point} and Tang\textit{et al.} \cite{tang2020searching} attempt to bridge the features of points and voxels by using point-to-voxel and voxel-to-point transformations in the backbone networks. Points provide detailed geometric information, while voxels are computationally efficient. PV-RCNN \cite{shi2023pv}, a two-stage 3D detector, incorporates the first-stage detector from Second \cite{yan2018second} and introduces the RoI-grid pooling operator for second-stage refinement. Qian \textit{et al.} \cite{qian2022badet} propose a lightweight region aggregation refine network (BANet) that constructs a local neighborhood graph to improve box boundary prediction accuracy. Additionally, Shi \textit{et al.} \cite{shi2023pv} propose PV-RCNN++, which utilizes VectorPool aggregation for better aggregating local point features with reduced resource consumption. Compared to pure voxel-based detection approaches, point-voxel based methods integrate features in the 3D backbone, allowing them to benefit from both fine-grained 3D shape and structure information obtained from points, resulting in improved detection accuracy, especially for two-stage detectors. However, this comes at the cost of increased inference time.

To ensure the inference speed of the model, our 3D backbone network retains a point-based design instead of using point-voxel-based approaches that integrate two types of features to gain benefits in the feature learning process. Instead, in the neck structure, we rely on the proposed point dilation mechanism to lift points to a new dimension and combine the detection head to jointly learn features of non-occupied space, thus ensuring the continuity of the model's receptive field. In summary, PDM-SSD provides a new approach to balance the inference speed and detection accuracy of the model.

\section{Method}
\label{sec:method}

\subsection{Overview}
\begin{figure}[t]
	\begin{center}
		\includegraphics[width=.8\textwidth]{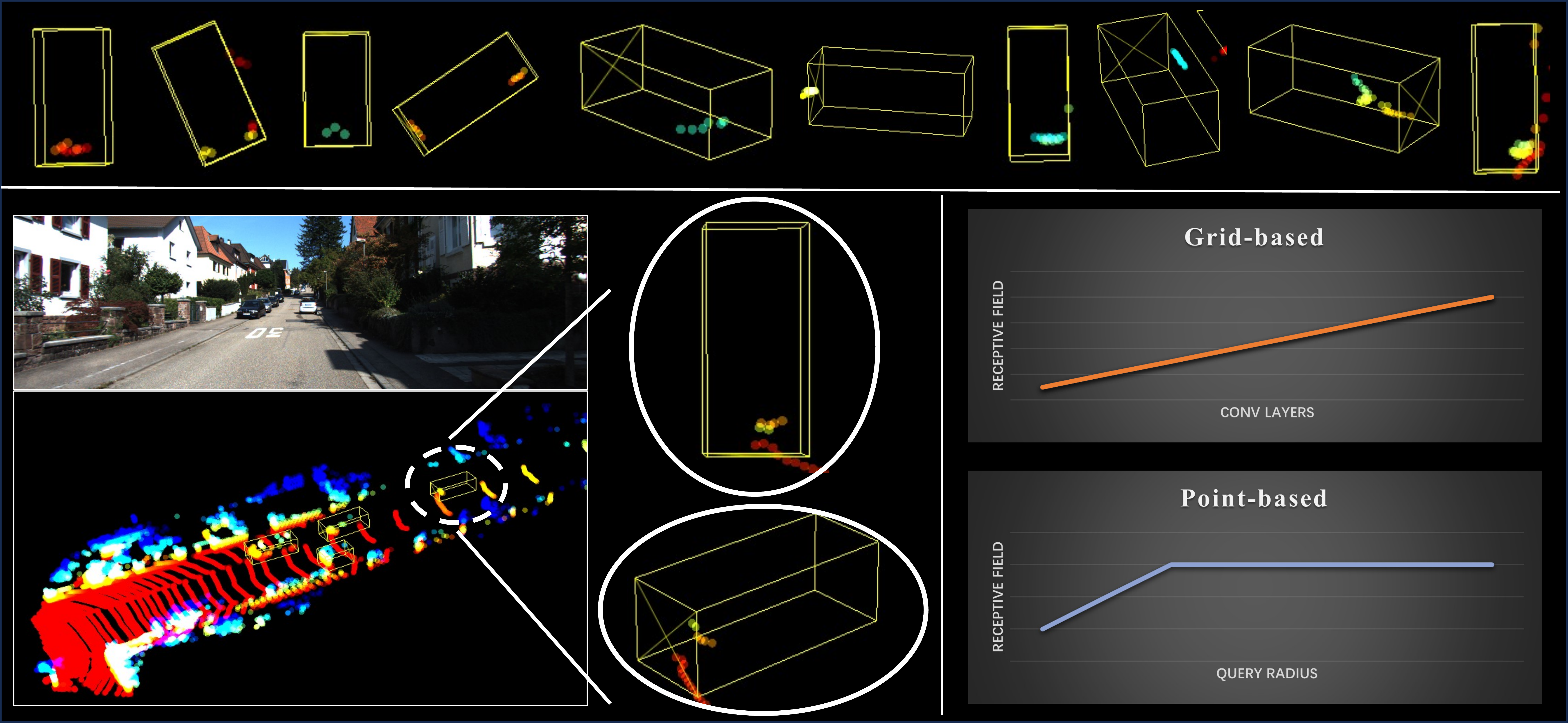}
	\end{center}
	\vspace{-0.3cm}
	\caption{Visualization on some very sparse and extremely incomplete targets on the KITTI dataset. For grid-based backbone networks, the grid continuously pads, convolves, and pools the operations, covering the space that the original point cloud does not occupy. The expansion of the receptive field is continuous and can better aggregate local features and combine features from different regions. Point-based methods can only extract features from existing points, and even if the number of surrounding points increases, the features remain unchanged. The receptive field is discontinuous and limited to local areas.}
	\label{fig3}
	\vspace{-0.3cm}
\end{figure}

As aforementioned, we aim to combine the advantages of grid-based and point-based detectors by addressing the issue of discontinuous receptive fields in current point-based models. To achieve this, we propose PDM-SSD, a novel and generic single-stage point-based 3D detector. The overall workflow of PDM-SSD is illustrated in Fig. \ref{fig2}, where the input LiDAR point clouds are first passed through the embedding network to expand the feature space of the points. Then, a PointNet-style 3D backbone network is employed to extract point-wise features. The 3D backbone network consists of several stages of downsampling modules, local feature aggregation modules, and multi-scale feature aggregation modules, ensuring that the sampled points learn rich geometric and semantic features. The neck network includes our proposed point dilation mechanism, where points are lifted to the grid level and feature filling is performed for the unoccupied space in the original point cloud using a special mechanism. The grid-wise features are used to regress the heatmap of the scene, providing global features of the objects through joint learning with the hybrid detection head.

In the following, we discuss in detail the implementation process of PDM-SSD. In Section \ref{sec:pf}, we examine the changes in the receptive fields of the grid-based and point-based models during the formulation learning process. In Sections \ref{sec:backbone}, \ref{sec:neck}, and \ref{sec:head}, we respectively introduce the specific structures of the PDM-SSD model's backbone, neck, and detection head. Finally, in Section \ref{sec:loss}, we explain the approach of joint learning in the model and the loss function.

\subsection{Problem formulation}
\label{sec:pf}
Let $P=\{p_n\}^N_{n=1}$ be a set of $N$ observed LiDAR points belonging to a scaene, where $p_n \in \mathbb{R}^{3}$ is a 3D point represented with spatial coordinates. Let $C$ be the centers location $C=\{c_m\}^M_{m=1}$ of the annotated ground-truth $M$ bounding boxes, $c_m=[c_{mx},~c_{my},~c_{mz},]\in \mathbb{R}^{3}$.
Due to the limitations of the resolution of the LiDAR, climate conditions, and occlusions, sparse and incomplete targets often appear in the scene, as shown in Fig. \ref{fig3}. Assuming the point cloud $P$ is voxelized into initial pillars $V=\{v_k\}^K_{k=1}$, where $K$ represents the number of pillars. Taking 2D sparse convolution as an example, after learning in the Grid-based detector with the convolutional layers stacked in the backbone network, the feature map can be simplified as:
\begin{equation}
	F^{g}_{i+1}=Pool(Conv(Padding(F^{g}_{i})))
\end{equation}
$i$ is the number of convolutional layers of the backbone. The padding layers in the network allocate the unoccupied space in the original point cloud, and the convolution layers and pooling layers fill these spaces with features. The receptive field of the current layer is:
\begin{equation}
	RF_{i+1}=RF_i+(k-1)\times S_i
\end{equation}
\begin{equation}
	S_i=\prod^{i}_{i=1}Sride_i
\end{equation}
Among them, $RF_{i+1}$ represents the receptive field of the current layer, $RF_i$ represents the receptive field of the previous layer, and $k$ represents the size of the convolution kernel. It can be seen that the receptive field of the model continuously increases, even learning features for the unoccupied space in the original point cloud. The feature maps of the Point-based backbone network can be represented as follows:
\begin{equation}
	\label{eq4}
	F^{p}_{i+1}=Pool(Group(Sampling(F^{p}_{i})))
\end{equation}
$i$ represents the stage of feature learning, $Sampling$ is the downsampling operation, $Group$ is the local feature aggregation operation, $Pool$ is the pooling layer, which is treated as a symmetric function to address the unordered nature of point clouds. Here, $F^{p}_{0}=E_{\theta}(P)$, where $E_{\theta}(\dots)$ is the feature augmentation network, usually a multi-layer perceptron or graph neural network, which expands the feature space of point clouds. As the number of learning stages increases, the query radius of the local feature extractor becomes larger to increase the receptive field of the model. However, the model only takes the original point cloud as input, and in the case where the target has points only in the local region, even if the query radius increases, the receptive field of the model remains the same, and the learning of local features still stays at the previous stage. Fig. \ref{fig3} illustrates this phenomenon intuitively. Under this problem, the model's ability to predict the semantics and geometry information of upsampled points on the target will decrease. The main purpose of PDM-SSD is to alleviate this problem.

\subsection{PointNet style 3D backbone}
\label{sec:backbone}
The currently popular point-based detectors, such as 3DSSD, IA-SSD, and DBQ-SSD, all use PointNet-style 3D backbones as the point semantic and geometric information extractors, and their advantages in detection accuracy and inference speed have been well demonstrated. In order to ensure lightweightness, our PDM-SSD also adopts this structure of 3D backbone. To highlight the performance gains after solving the problem of discontinuous receptive fields, we do not make too many changes to the backbone, following the design of IA-SSD in general. The specific structure is shown in Fig. \ref{fig2}.

The point cloud is first passed through a vanilla feature augmentation network stacked with MLP to increase the feature dimension. Then, it enters a feature extractor consisting of four repeated SA (Set Abstraction) modules to learn geometric and semantic information, as shown in Eq. \ref{eq4}. The point-wise features outputted at each stage are denoted as $\{F^p_{1}, F^p_{2}, F^p_{3}, F^p_{4}\}$. Specifically, before each stage, we downsample the point cloud to reduce the model's spatial complexity and feature redundancy. Then, we use PointNet for local feature extraction, which can be divided into three steps: 1) Point indexing: using the current stage sampled points as centers, perform vanilla ball query within the range of the sampled points in the previous stage to index the $k$ nearest points to each center within a certain radius $r$. 2) Feature learning: use MLP to learn the features of the points within the ball, further improving the learning depth. 3) Pooling: perform max-pooling operation on the features of the points within the ball in the feature dimension. Due to the unordered nature of point clouds, this operation ensures that even if the order of the point cloud is changed, the pooled features remain unchanged. In each stage, we simultaneously use two sets of combinations of radius and sampling number for multi-scale feature extraction, and aggregate the multi-scale features after the pooling layer. In summary, the point quantities of $F^p_{1}, F^p_{2}, F^p_{3}, F^p_{4}$ decrease while the feature dimensions gradually increase.

It should be noted that in the model, $F^p_{1}, F^p_{2}$ adopt Farthest Point Sampling (FPS) downsampling method, while $F^p_{3}, F^p_{4}$ adopt foreground point downsampling method with semantic embedding, following the approach of IA-SSD. The former ensures the global coverage of sampling points in the presence of a large number of redundant points, reducing global information loss. The latter ensures a high recall rate of foreground points, reducing target information loss. The specific implementation is shown in Fig. \ref{fig2}. We add a network branch $S(\cdot)$ to the SA module in the second and third stages to extract the semantic information of sampling points. Then, the sampling points in the next stage are selected from the points with the highest probability of being foreground points, as shown in the following equation:
\begin{equation}
	sample\_index_{i+1}=topK(S(F_i)) \quad i=2,3
\end{equation}
From this, we obtain a small number of foreground sampling points with their rich local geometric and semantic information in point-wise features.

\subsection{Neck with Point Dilation Mechanism}
\label{sec:neck}
\begin{figure}[t]
	\begin{center}
		\includegraphics[width=.5\textwidth]{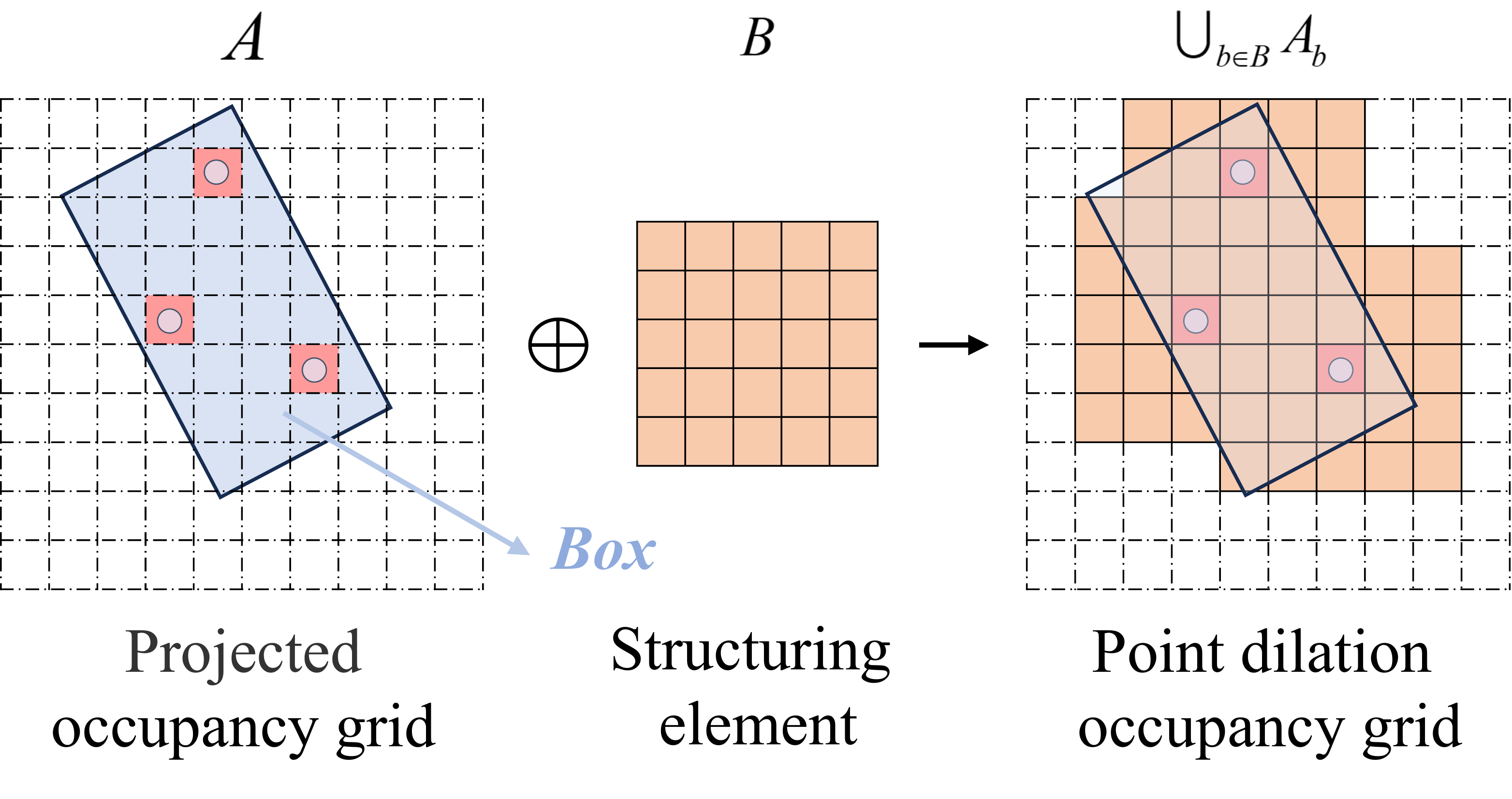}
	\end{center}
	\vspace{-0.3cm}
	\caption{Point dilation operation. The point cloud is first projected onto a 2D binary occupancy grid and then dilated with a structural element. The new feature map covers many areas that were not occupied by the original point cloud, especially the region where the target box is located (blue box). The feature at the center position is of great interest to the detector.}
	\label{fig4}
	\vspace{-0.3cm}
\end{figure}

The point-wise receptive field is still limited to the space occupied by the original point cloud until \ref{sec:backbone}. In order to expand the learning scope of the model, we propose the Point Dilation Mechanism, which consists of two steps: point dilation and feature filling.

\begin{figure}[t]
	\begin{center}
		\includegraphics[width=.5\textwidth]{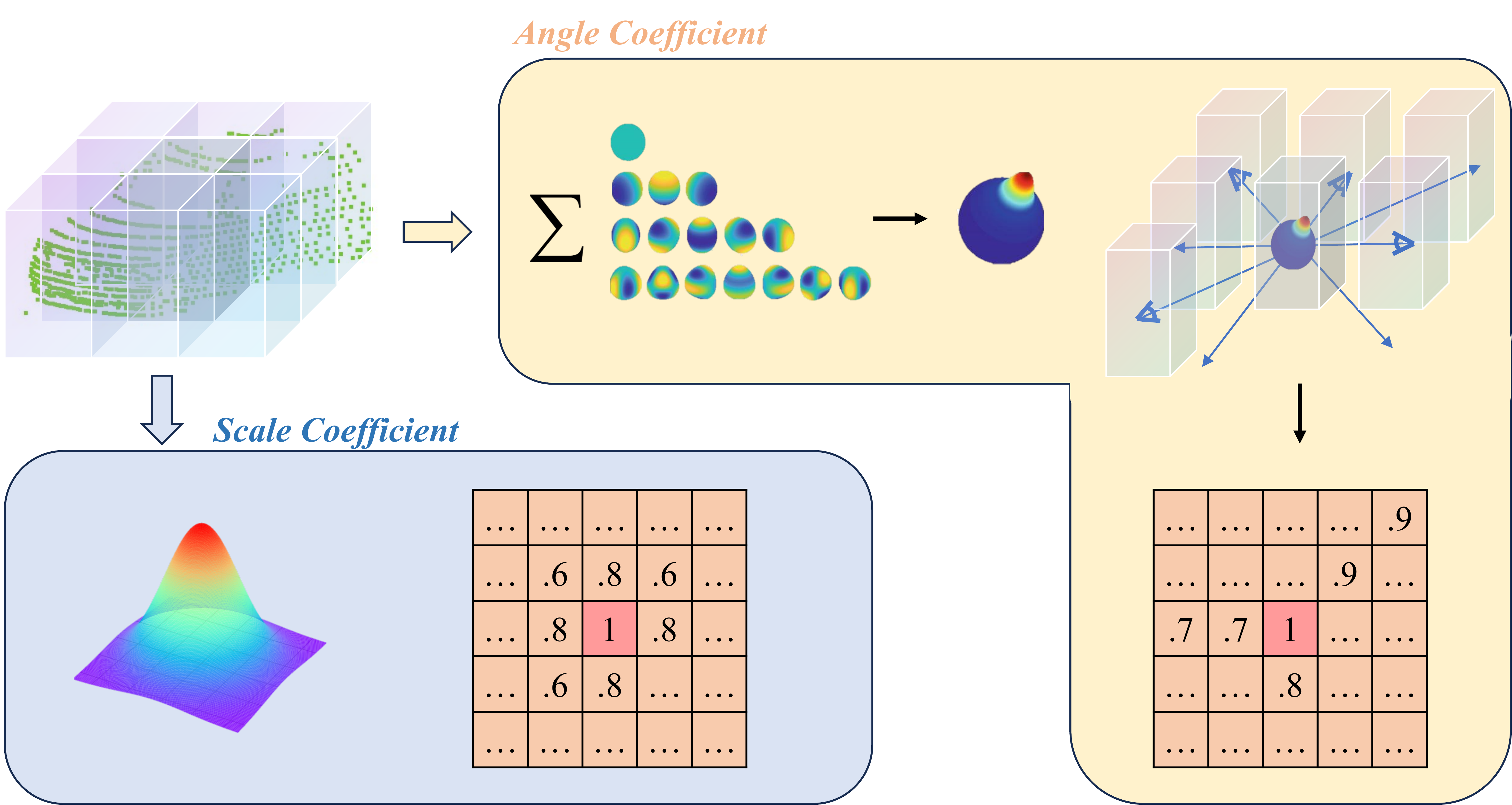}
	\end{center}
	\vspace{-0.3cm}
	\caption{Feature Filling operation. We propose a feature filling method based on spatial separation coefficient. We use point-wise feature learning for Angle Coefficient and Scale Coefficient. The former is achieved by the superposition of spherical harmonics, while the latter is achieved by Gaussian probability density function. The new feature is the weighted sum of the inflated center feature and these two coefficients.}
	\label{fig5}
	\vspace{-0.3cm}
\end{figure}

\textbf{Point Dilation (PD)}. Dilation (usually represented by $\oplus$) is one of the basic operations in mathematical morphology. Image dilation is a commonly used morphological processing algorithm in the field of image processing, which utilizes a structuring element to probe and expand the shapes present in the input image, aiming to connect connected regions or eliminate noise \cite{wu2010morphological}. Our PD method borrows the idea of image dilation to achieve spatial expansion. Specifically, for $F_4^p$ obtained from \ref{sec:backbone}, we use Eq. \ref{eq6} to sparsely project it onto a grid, resulting in $F_4^g$.
\begin{equation}
	\label{eq6}
	F_i^g=G(F_i^p, W,H,\epsilon)
\end{equation}
Where $G(\cdot)$ is the projection function, $W$ and $H$ are the width and height of the feature map respectively, and $\epsilon$ is another hyperparameter representing the spatial range of the computed point cloud. It is worth noting that in order to maintain unique indices for the sparse grid, for the sampled points that are projected onto the same grid, we sum their features along the feature dimension to obtain the feature of the current cell. At this point, the feature map is very sparse. We set the occupied cells to 1 and binarize the feature map, and then perform dilation operation:
\begin{equation}
	F_i^g\oplus B=\cup_{b\in B} {F_i^g}_b
\end{equation}
Where $B$ is a structuring element, and we use a $5\times 5$ matrix consisting entirely of 1s. This process is illustrated in Fig. \ref{fig5}, and the dilated binary feature map contains a significant amount of unoccupied space in the original point cloud.

\textbf{Feature Filling (FF)}. Although PD expands the occupancy space of the feature map, there are no learnable features available on these new cells, which presents a significant challenge in filling the features. We believe that the filled features must adhere to several principles: 1) Learnability: In theory, unoccupied cells and occupied cells should have equal importance in providing information, especially for the position of the target area. We must ensure that these cells have sufficient learning depth and flexibility. 2) Spatial correlation: Previous works such as 3DSSD and IA-SSD have shown that point-wise features at this stage can already regress the rough center position of the target. Therefore, the newly filled features should have a certain correlation with the dilation center to prevent feature fragmentation and preserve the predictive ability of the original dilation center. 3) Cross-correlation: It is evident that multiple dilation centers may affect the features of the same cell. The current cell should have the ability to connect multiple dilation centers to achieve multi-center interaction and learn features with a larger receptive field. To address this, we propose a feature learning method based on spatial separation coefficients. Specifically, we consider the angle and scale aspects and fill the angle and scale influence coefficients for the structuring element $B$. The features of the newly filled cell are then obtained by weighting the dilation center with the separation coefficients. For cells affected by multiple centers, a simple height compression is used to aggregate the features.

\textbf{Angle Coefficient (AC) $\alpha$.} The perception of objects in 3d space by humans or machines is influenced by the observation angle and viewpoint. Taking ray tracing as an example, the color of a certain grid in space varies when observed from different angles. We extend this consensus to feature space, where the features in the surrounding space are related to the angle between them and the expansion center. The conditions of learnability and angle correlation naturally lead us to think of the commonly used spherical harmonics coefficients in simple lighting descriptions.
\begin{equation}
	\label{eq8}
	\nabla^2 f=\frac{1}{r^2}\frac{\partial}{\partial r}(r^2\frac{\partial f}{\partial r}) + \frac{1}{r^2 \sin{\theta}} \frac{\partial}{\partial \theta} (\sin{\theta} \frac{\partial f}{\partial \theta}) + \frac{1}{r^2 \sin{\theta}^2}\frac{\partial^2 f}{\partial \varphi
		^2}=0
\end{equation}
The spherical harmonics are the angular part of the solution to the Laplace equation in spherical coordinates. The Laplace equation in spherical coordinates can be written as Eq. \ref{eq8}. Using the method of separation of variables, we can assume that $f(r,\theta,\varphi)=R(r)Y(\theta,\varphi)=R(r)\Theta(\theta)\Phi(\varphi)$. Here, $Y(\theta,\varphi)$ represents the angular part of the solution, which is also known as the spherical harmonics. More intuitively, the spherical harmonics can also be expressed as:
\begin{equation}
	\label{eq9}
	Y^m_l(\theta,\varphi)=(-1)^m\sqrt{\frac{(2l+1)}{4\pi}\frac{(l-|m|)!}{(l+|m|)!}}P^m_l(\cos{\theta})\exp(im\varphi)
\end{equation}
\begin{equation}
	\label{eq10}
	P^m_l(x)=(1-x^2)^{\frac{|m|}{2}}\frac{d^{|m|}}{dx^{|m|}}P_l(x)
\end{equation}
\begin{equation}
	\label{eq11}
	P_l(x)=\frac{1}{2^l l!}\frac{d^l}{dx^l}(x^2-1)^l
\end{equation}
The spherical harmonic function is only dependent on angles, where $l$ and $m$ are the degree index and order of the associated Legendre polynomial $P^m_l$. In computer graphics, the spherical harmonic function is similar to the Fourier transform in representing lighting. The result is obtained by weighting multiple spherical harmonic coefficients with the basis of spherical harmonic functions. The more spherical harmonic coefficients are used, the stronger the expressive power and the closer it is to the original function. In this method, we treat the dilation center as a sphere and calculate the angular coefficients of the new cell on the structuring element $B$ based on the spherical harmonic coefficients. Specifically, we use a prediction network $SH_\theta(\cdot)$ to regress the 16 spherical harmonic coefficients of the fourth order in $F^g_4$ for each dilation center. Then, the weighted calculation result is used as the angular coefficients of the new cell, as shown in Eq. \ref{eq12}.
\begin{equation}
	\label{eq12}
	\alpha=\sum_{l=0} \sum_{m=-1}^{l+1} c_l^m Y^m_l(\theta,\varphi)
\end{equation}
The input of Eq. \ref{eq12} are the angle between the center point of the newly inflated cell and the center of the cell where the inflation occurs.

\textbf{Scale Coefficient (SC) $\beta$.} As mentioned above, the current point-wise feature $F^g_4$ already contains certain geometric information, so the new cell features around it are scale-dependent on the dilation center. Similar to AC, SC is also reflected by the values filled in the structuring element $B$. Specifically, we use a Gaussian density function to calculate SC, taking the relative position between the dilated new cell and the cell where the dilation center is located, as well as the learned variance, as inputs. The formula is as follows:
\begin{equation}
	\label{eq13}
	\beta= G(x,\mu,\Sigma)
\end{equation}
\begin{equation}
	\label{eq14}
	G(x,\mu,\Sigma)=G(x_1,x_2,\dots,x_D,\mu,\Sigma)=\frac{1}{(2\pi)^{D/2}{\vert\Sigma\vert}^{1/2}} exp(-\frac{1}{2}(x-\mu)^T{\Sigma}^{-1}(x-\mu))
\end{equation}

\begin{figure}[t]
	\begin{center}
		\includegraphics[width=.5\textwidth]{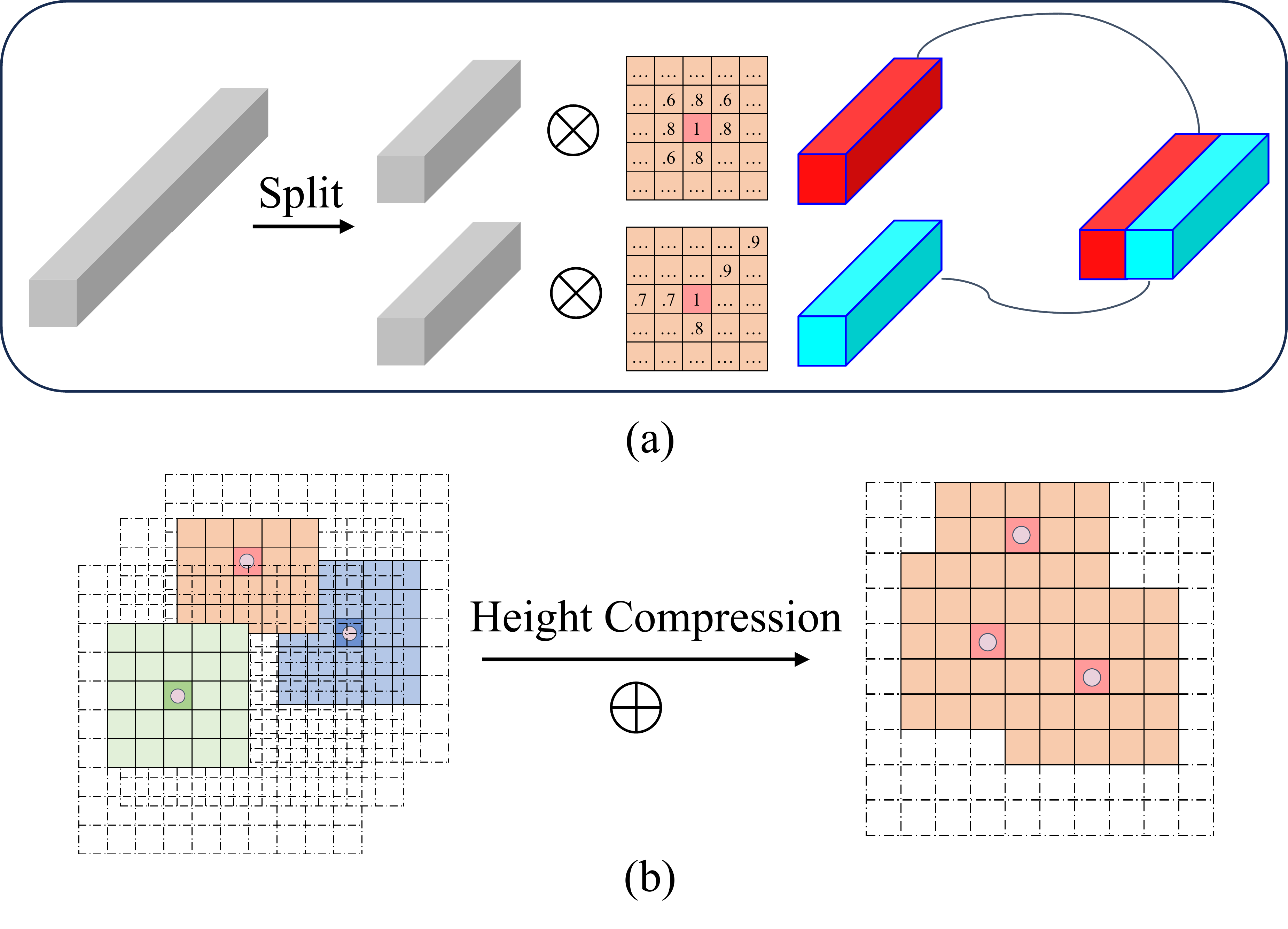}
	\end{center}
	\vspace{-0.3cm}
	\caption{(a) Coefficients fusion. In order to maintain the nonlinearity of different cell features, we first decompose the point-wise features and weight them separately using AC and SC. The final feature is the sum of these two parts. (b) Height compression. For cells that are influenced by multiple dilation centers, we directly add their multiple features together, retaining the effect of each dilation center.}
	\label{fig6}
	\vspace{-0.3cm}
\end{figure}

In this method, $G(x, \mu, \Sigma)$ represents independent and identically distributed bivariate Gaussians, where $\mu$ is the position of the inflated center cell and $x$ is the position of the new cell. $\Sigma=\sigma^2E$ is a scaled identity matrix. We also add a scale prediction branch network $S_\theta(\cdot)$ to regress the variance of the inflated center, and then multiply it by a 2D identity matrix $E$ to calculate the final scale coefficient using Eq. \ref{eq13}.

\textbf{Coefficients Fusion.} The new features filled during the dilation process with angle and scale coefficients have a direct linear relationship with the dilation center, which is detrimental to the robustness of deep models. To address this, we adopt a channel-splitting approach for coefficients fusion, as shown in Fig. \ref{fig6} (a). Specifically, we split the point-wise features of the dilation center into two parts in the depth dimension, and multiply them with the angle and scale coefficients respectively. Finally, we sum them up to obtain the new point-wise features. Through this operation, the relationship variables between the new features and the dilation center increase to binary, maintaining the non-linearity of the new features. For cells influenced by multiple dilation centers, we want to preserve the effects of all dilation centers to connect the surrounding features. We achieve this by using a height compression operation as shown in Fig. \ref{fig6} (b), where the effects of all active dilation centers at that location are stacked, reducing the spatial complexity of the training and inference processes. Thus, we obtain a relatively dense 2D feature map.

\subsection{Hybrid Head}
\label{sec:head}
\begin{figure}[t]
	\begin{center}
		\includegraphics[width=.7\textwidth]{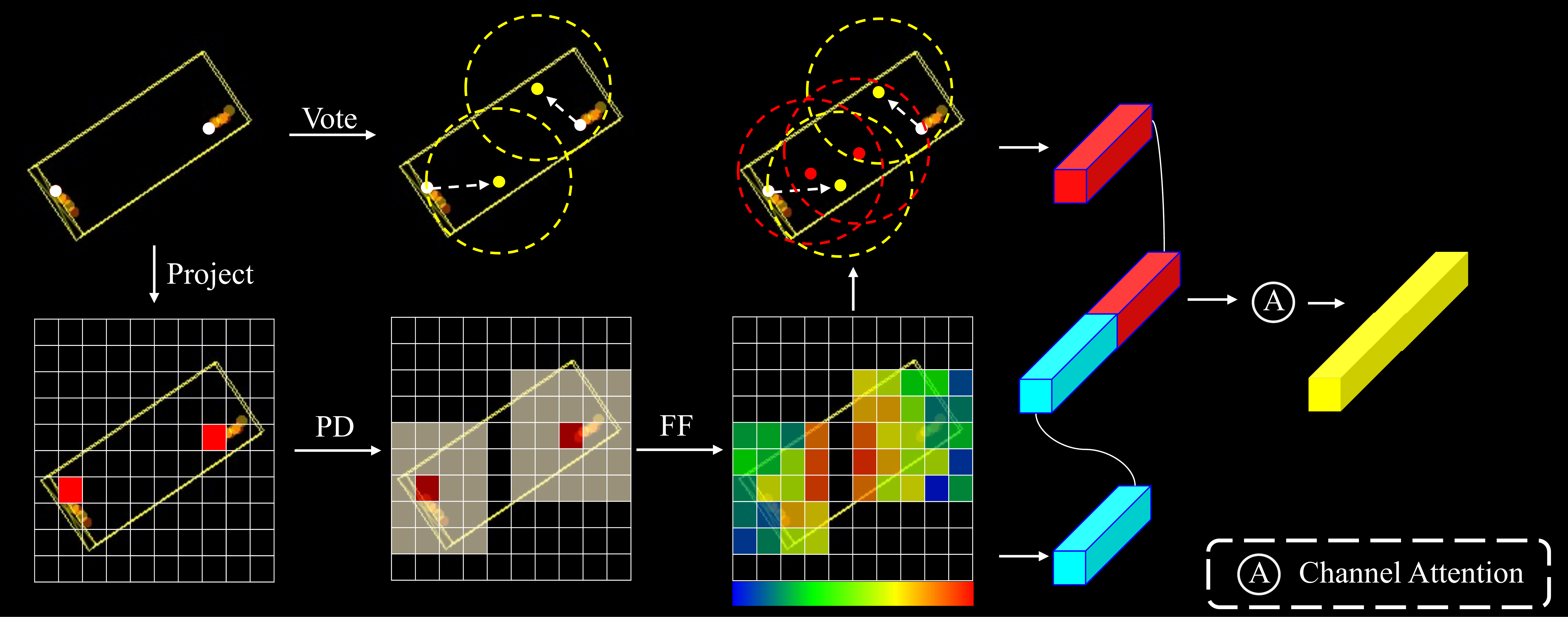}
	\end{center}
	\vspace{-0.3cm}
	\caption{Joint Learning. In the case of sparsity and extremely incomplete targets, on one hand, the vote points generated by the vote network may deviate from the target center, and on the other hand, the point-wise features have a prediction for the target category lower than the threshold. We use the auxiliary information generated by the neck to improve these issues. First, we supplement the vote point set with the learned heatmap. Then, we concatenate the point-wise features and grid features, and pass them through the channel attention network to form new features for predicting the parameters of the target box.}
	\label{fig7}
	\vspace{-0.3cm}
\end{figure}

We propose a hybrid detection head to simultaneously extract point-wise features and grid features, as shown in Fig. \ref{fig7}. For the point-wise features $F^p_4$, inspired by VoteNet, they are first input into a voting network to move the sampled points towards the center of the objects as much as possible. Then, the context features of the objects are extracted using the moved positions as centers. Finally, these features are used for semantic classification and regression of geometric parameters. This part follows the basic structure of a traditional point-based detection head. However, as described in Section \ref{sec:pf}, due to the limited receptive field, the information of objects in the scene is not fully explored, especially for sparse and locally scattered objects, where their voting centers may be inaccurate or the context features may have a probability of being classified as objects below the certain threshold.

The grid feature, with the support of angle and scale coefficients, not only expands the receptive field of point-wise features but also enhances the connectivity of features from multiple inflated centers, potentially enabling the extraction of holistic features from sparse and incomplete targets. To achieve this goal, we construct a sparse heatmap using annotation information to supervise the learning of grid features, following the design of the dense CenterHead in the training process. Additionally, we find that even treating this part of the network as an auxiliary learning task can significantly improve the performance of the point-based detection head. Therefore, we design the model from the perspectives of both auxiliary learning and joint learning.

\textbf{Auxiliary Learning (AL).} The $F^g_4$ used for heatmap regression is essentially obtained by dilating $F^p_4$ used for prediction. We believe that supervising $F^g_4$ with sparse heatmaps will also increase the receptive field of $F^p_4$, thereby obtaining more accurate detection results. In addition, another advantage of auxiliary learning is that the auxiliary network does not participate in prediction and inference stages of the model, thus preserving the inference speed of the point-based detector completely. We will demonstrate its superiority in the experimental section.

\textbf{Joint Learning (JL).} In joint learning, we will fully utilize grid features and learned scene heatmaps. As shown in Fig. \ref{fig7}, in addition to the votes generated by the vote network from point-wise features, we supplement the centers of the top $K$ maximum values in the predicted heatmap. These two sets of points are aggregated together to learn contextual features through the point-based head. Then, we contact the contextual point-wise features with the grid features of the cells where these points are located in the depth dimension. Finally, after a simple channel attention, the features are used for semantic classification and detection box parameter regression. The above operations can not only supplement the spatial position information of the scene targets provided by the heatmap, but also adaptively fuse point-wise features and grid features, thereby compensating for the inaccuracies caused by the limited receptive field of the point-based detector and the low semantic probabilities of the targets. We will explain this in detail in \ref{sec:pdm}.

\subsection{End-to-End Learning}
\label{sec:loss}
The proposed PDM-SSD can be trained in an end-to-end fashion. In the 3D backbone, the sampling loss is calculated.
\begin{equation}
	L_{sample}=-\sum_{c=1}^C{(Mask_i \cdot CELoss_i)}
\end{equation}
\begin{equation}
	CELoss_i = s_i \log(\hat{s_i}) + (1-s_i)\log{1-\hat{s_i}}
\end{equation}
\begin{equation}
	Mask_i=\sqrt[3]{\frac{\min{f^*,b^*}}{\max{f^*,b^*}}\times \frac{\min{l^*,r^*}}{\max{l^*,r^*}} \times \frac{\min{u^*,d^*}}{\max{u^*,d^*}}}
\end{equation}
$CELoss$ is the cross-entropy loss for predicting the category of sampled points, which first appeared in IA-SSD. It greatly improves the recall rate of foreground points in sampled points. $Mask$ is the weight for the centrality of sampled points. $f^*,b^*,l^*,r^*,u^*,d^*$ represent the distances between the sampled points and the six faces of the target box. We borrowed the design from 3DSSD, which believes that points closer to the center are more conducive to accurately regressing semantic categories and geometric parameters. By multiplying the centrality with the cross-entropy loss, the closer the sampled points are to the center, the greater the loss will be. The model will prioritize improving the foreground probability $s_i$ of these points, so that during inference, the model will prioritize selecting these points. Although SPSNet has shown that points closer to the center are not necessarily better, it is still better than methods that sample foreground points equally. Besides, SPSNet requires extra time to learn the stability of points.

The loss in the detection head is divided into two parts: the point-based loss $L_{p}$ and the grid-based heatmap prediction loss $L_{heatmap}$. The former consists of three components: the loss of the vote points $L_{vote}$, the loss of point semantics prediction $L_{cls}$, and the loss of target box geometric parameter regression $L_{reg}$. Additionally, we also include a regularization loss in the training process of PDM-SSD.
\begin{equation}
	L_{all}=L_{sample}+L_{p}+L_{heatmap}+L_2
\end{equation}
\begin{equation}
	L_{p}=L_{vote}+L_{cls}+L_{reg}
\end{equation}
In particular, the box generation loss can be further decomposed into location, size, angle-bin, angle-res, and corner parts:
\begin{equation}
	L_{reg}=L_{loc}+L_{size}+L_{angle-bin} + L_{angle-res} + L_{corner}
\end{equation}
All these losses will be jointly optimized using a multi-task learning approach.

\section{Experiments}
\label{sec:experiments}
In this section, we will provide detailed experimental results to demonstrate the efficiency and accuracy of PDM-SSD. Specifically, we introduced the specific settings and implementation details of the experiments in Section \ref{sec:setup}. Then, the comparison results between PDM-SSD and current state-of-the-art methods were reported in Section \ref{sec:comparison}. Following that, in Section \ref{sec:ablation}, a ablation study was conducted to demonstrate the rationality of the model design. Furthermore, the inference efficiency of PDM-SSD was analyzed in Section \ref{sec:runtime}. Finally, in Section \ref{sec:pdm}, we provided a large number of instances to demonstrate the superiority of PDM in sparse and incomplete object detection.

\subsection{Setup}
\label{sec:setup}

\begin{table}
	\centering
	\caption{Quantitative comparison with state-of-the-art methods on the KITTI \textit{test} set for \textit{Car} BEV and 3D detection, under the evaluation metric of 3D Average Precision ($AP$) of 40 sampling recall points. The best and our PDM-SSD results are highlighted in \textbf{bold} and \uline{underlined}, respectively}
	\resizebox{\textwidth}{!}{
	\begin{tabular}{c|c|c|ccc|ccc} 
		\toprule[0.75mm]
		\multirow{3}{*}{Method} & \multirow{3}{*}{Structure} & \multirow{3}{*}{Type} & \multicolumn{3}{c|}{$AP_{3D}@Car$(IoU=0.7)}                 & \multicolumn{3}{c}{$AP_{BEV}@Car$(IoU=0.7)}                  \\ 
		\cline{4-9}
		&                            &                       & \multicolumn{3}{c|}{R40}                         & \multicolumn{3}{c}{R40}                           \\
		&                            &                       & Easy           & Moderate       & Hard           & Easy           & Moderate       & Hard            \\ 
		\hline
		VoxelNet \cite{zhou_voxelnet_2017}                & Voxel-based                & 1-stage               & 77.47          & 65.11          & 57.73          & 87.95          & 78.39          & 71.29           \\
		PointPillars \cite{lang_pointpillars_2019}            & Voxel-based                & 1-stage               & 82.58          & 74.31          & 68.99          & 90.07          & 86.56          & 82.81           \\
		SECOND \cite{yan2018second}                  & Voxel-based                & 1-stage               & 84.65          & 75.96          & 68.71          & 89.39          & 83.77          & 78.59           \\
		3DIoULoss \cite{zhou2019iou}               & Voxel-based                & 2-stage               & 86.16          & 76.5           & 71.39          & 90.23          & 86.61          & 86.37           \\
		TANet \cite{liu2020tanet}                   & Voxel-based                & 1-stage               & 84.39          & 75.94          & 68.82          & 91.58          & 86.54          & 81.19           \\
		Part-A2 \cite{shi2020points}                 & Voxel-based                & 2-stage               & 87.81          & 78.49          & 73.51          & 91.7           & 87.79          & 84.61           \\
		CIASSD \cite{zheng2021cia}                  & Voxel-based                & 1-stage               & 89.59          & 80.28          & 72.87          & 93.74          & 89.84          & 82.39           \\
		SASSD \cite{he2020structure}                   & Voxel-based                & 1-stage               & 88.75          & 79.79          & 74.61          & 93.74          & 89.84          & 82.39           \\
		Associate-3Det \cite{du2020associate}          & Voxel-based                & 1-stage               & 85.99          & 77.4           & 70.53          & 91.4           & 88.09          & 82.96           \\
		SVGA-Net \cite{he2022svga}                & Voxel-based                & 1-stage               & 87.33          & 80.47          & 75.91          & -              & -              & -               \\ 
		\hline
		Fast Point R-CNN \cite{chen2019fast}        & Point-Voxel                & 2-stage               & 85.29          & 77.4           & 70.24          & 90.87          & 87.84          & 80.52           \\
		STD \cite{yang2019std}                     & Point-Voxel                & 2-stage               & 87.92          & 79.71          & 75.09          & -              & -              & -               \\
		PV-RCNN \cite{shi2020pv}                 & Point-Voxel                & 2-stage               & \uline{90.25}  & 81.43          & 76.82          & \uline{94.98}  & \uline{90.62}  & 86.14           \\
		EQ-PVRCNN \cite{yang2022unified}               & Point-Voxel                & 2-stage               & 90.13          & \uline{82.01}  & \uline{77.53}  & 94.55          & 89.09          & \uline{86.42}   \\
		VIC-Net \cite{jiang2021vic}                 & Point-Voxel                & 1-stage               & 88.25          & 80.61          & 75.83          & -              & -              & -               \\
		HVPR \cite{noh2021hvpr}                    & Point-Voxel                & 1-stage               & 86.38          & 77.92          & 73.04          & -              & -              & -               \\ 
		\hline
		PointRCNN \cite{shi2019pointrcnn}               & Point-based                & 2-stage               & 86.96          & 75.64          & 70.7           & 92.13          & 87.39          & 82.72           \\
		3D IoU-Net \cite{li20203d}              & Point-based                & 2-stage               & 87.96          & 79.03          & 72.78          & 94.76          & 88.38          & 81.93           \\
		3DSSD \cite{yang20203dssd}                   & Point-based                & 1-stage               & 88.36          & 79.57          & 74.55          & 92.66          & 89.02          & 85.86           \\
		IA-SSD \cite{zhang2022not}                  & Point-based                & 1-stage               & 88.34          & 80.13          & 75.04          & 92.79          & 89.333         & 84.35           \\
		DBQ-SSD \cite{yang2022dbq}                 & Point-based                & 1-stage               & 87.93          & 79.39          & 74.4           & -              & -              & -               \\
		\textbf{PDM-SSD}                 & Point-based                & 1-stage               & \textbf{88.74} & \textbf{80.87} & \textbf{75.63} & \textbf{93.07} & \textbf{89.92} & \textbf{85.12}  \\
		\bottomrule
	\end{tabular} }
	\label{tabel1}
\end{table}

\textbf{Benchmark datasets.} The KITTI dataset is a dataset sponsored by the Karlsruhe Institute of Technology and the Toyota Technological Institute at Chicago for research in the field of autonomous driving. The widely-used dataset contains 7481 training samples with annotations in the camera field of vision and 7518 testing samples. Following the common protocol, we further divide the training samples into a training set (3,712 samples) and a validation set (3,769 samples). Additionally, the samples are divided into three difficulty levels: simple, moderate, and hard based on the occlusion level, visibility, and bounding box size. The moderate average precision is the official ranking metric for both 3D and BEV detection on the KITTI website.

\textbf{Evaluation metrics.} To provide a comprehensive performance evaluation, we evaluated our PDM-SSD on both the KITTI 3D and BEV object detection benchmarks. Generally, average precision ($AP$) based on Intersection over Union (IoU) is commonly used for both the 3D and BEV tasks. The experiments primarily focused on the commonly-used \textit{Car} category and were evaluated using the average precision metric with an IoU threshold of 0.7. To ensure an objective comparison, we utilized both the $AP$ with 40 recall points ($AP_{40}$) and the {AP} with 11 recall points ($AP_{11}$). The 3D-NMS threshold for metric calculation was set at 0.1, and the object score threshold was set at 0.1.

\textbf{Training details.} To ensure fair comparison, the training parameters of PDM-SSD are kept consistent with IA-SSD. Specifically, we use the Adam optimizer with $\beta_1=0.9$ and $\beta_2=0.85$ to optimize PDM-SSD. The weight decay coefficient is set to 0.01, and the momentum coefficient is set to 0.9. The model is trained for 80 epochs with a batch size of 16 on a Nvidia A40 GPU. The initial learning rate is set to 0.01, which is decayed by 0.1 at 35 and 45 epochs and updated with the one cycle policy. We initialized the weights of the heatmap prediction network in the neck module with values generated from a normal distribution. The training range of point clouds in the KITTI dataset is $[0,-40,-3,70.4,40,1]$, corresponding to $[x_{min},y_{min},z_{min},x_{max},y_{max},z_{max}]$. The grid size in the neck module is $176\times 200$, the size of the structural element is $5\times 5$, and the scale of the spherical harmonic coefficients is 3. The hybrid head predicted heatmaps and added the top 256 points with the highest foreground probabilities to the vote point set. We applied common scene-level data augmentation strategies to enhance the robustness of the model, including: 1) randomly rotating the scene along the $z$ axis within the range of $[-4/\pi,4/\pi]$ with a probability of 50\%; 2) randomly flipping the scene along the $x-z$ plane; 3) randomly scaling the scene within the range of $[0.95,1.05]$. Moreover, a sufficient number of targets was necessary for PDM-SSD to learn a more complete feature distribution. To achieve this, we employed object-level data augmentation methods to transform objects from other scenes. Specifically, 15 cars were copied to the current scene.

\textbf{Base detector.} Currently, the well-performing point-based detectors are all PointNet-style detectors, and their structures are quite similar, with the main difference lying in the different downsampling methods. PDM-SSD takes IA-SSD, which performs the best in terms of accuracy and inference speed on KITTI, as the base detector. In the experiments, the backbone of PDM-SSD used for auxiliary training is exactly the same as IA-SSD, making it very intuitive to see the advantages of PDM-SSD. In Section \ref{sec:pdm}, we will also provide a detailed comparison between PDM-SSD and IA-SSD at the object level.

\begin{figure}
	\centering
	\includegraphics[width=1\linewidth]{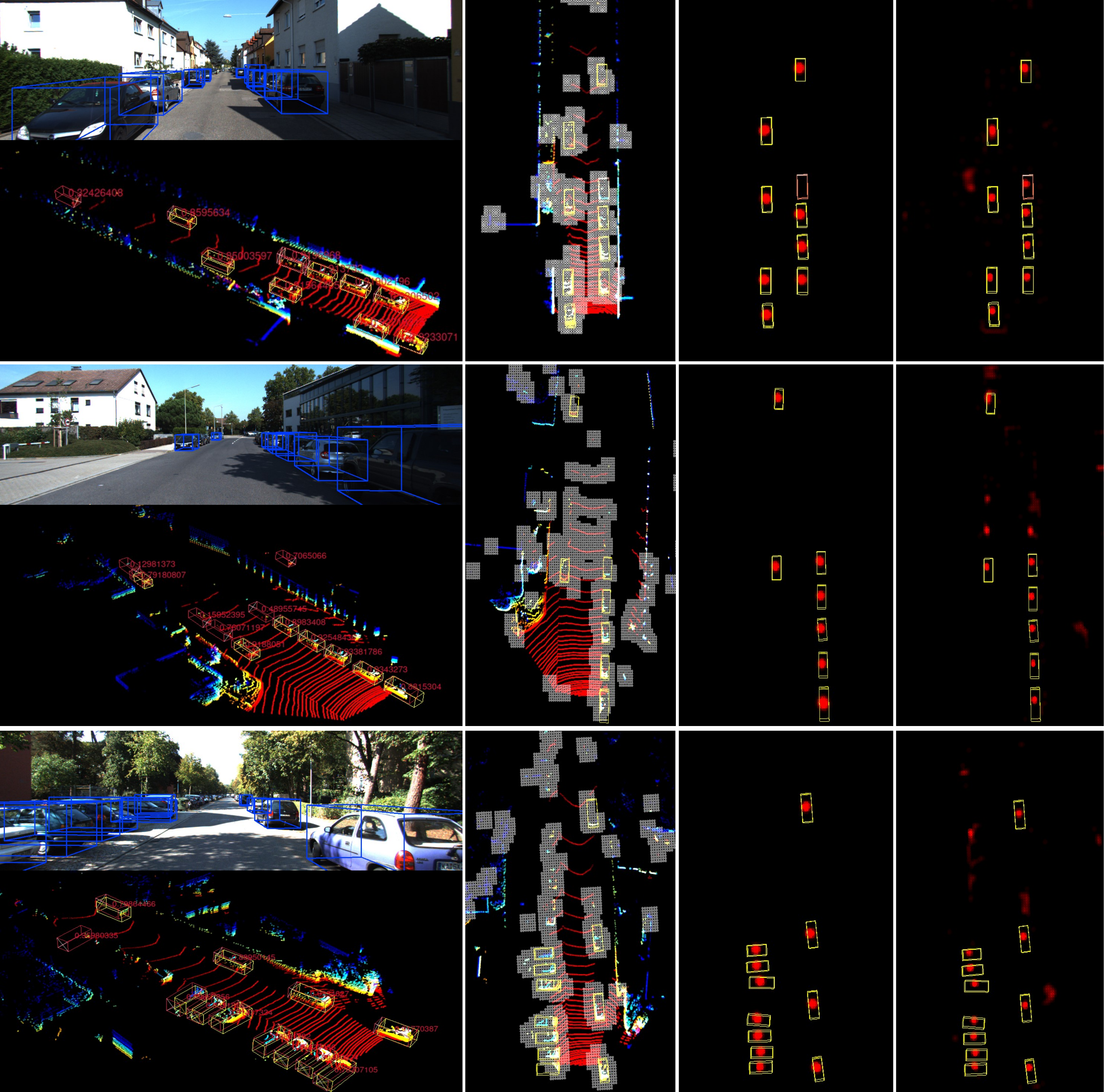}
	\caption{The detection results of PDM-SSD on a subset of KITTI \textit{val} set samples. From left to right, respectively: image and point clouds with predicted bounding boxes, space covered by grids after point dilatation, ground truth of space coverage heatmap in the scene, and predicted values of space coverage heatmap in the scene.}
	\label{fig:detectiveresult}
\end{figure}

\subsection{Comparison with State-of-the-Arts}
\label{sec:comparison}

\begin{table}
	\centering
	\caption{Quantitative comparison with state-of-the-art methods on the KITTI \textit{val} set for \textit{car} BEV and 3D detection, under the evaluation metric of 3D Average Precision (AP) of 11 and 40 sampling recall points. The best and our PDM-SSD results are highlighted in \textbf{bold} and \uline{underlined}, respectively}
	\resizebox{\textwidth}{!}{
	\begin{tabular}{c|c|c|ccc|ccc|ccc|ccc} 
		\toprule[0.75mm]
		\multirow{3}{*}{Method} & \multirow{3}{*}{Structure} & \multirow{3}{*}{Type} & \multicolumn{6}{c|}{$AP_{3D}@Car$(IoU=0.7)}                                                                & \multicolumn{6}{c}{$AP_{BEV}@Car$(IoU=0.7)}                                                                             \\ 
		\cline{4-15}
		&                            &                       & \multicolumn{3}{c|}{R11}                         & \multicolumn{3}{c|}{R40}                                 & \multicolumn{3}{c|}{R11}                                 & \multicolumn{3}{c}{R40}                           \\
		&                            &                       & Easy           & Moderate       & Hard           & Easy           & Moderate               & Hard           & Easy                   & Moderate       & Hard           & Easy           & Moderate       & Hard            \\
		\hline
		VoxelNet* \cite{zhou_voxelnet_2017}               & Voxel-based                & 1-stage               & 81.97          & 65.46          & 62.85          & -              & -                      & -              & 89.6                   & 84.81          & 78.57          & -              & -              & -               \\
		PointPillars* \cite{lang_pointpillars_2019}           & Voxel-based                & 1-stage               & 86.44          & 77.28          & 74.65          & 87.75          & 78.38                  & 75.18          & 89.66                  & 87.16          & 84.39          & 92.05          & 88.05          & 86.67           \\
		SECOND* \cite{yan2018second}                 & Voxel-based                & 1-stage               & 88.61          & 78.62          & 77.22          & 90.55          & 81.61                  & 78.61          & 90.01                  & 87.92          & 86.45          & 92.42          & 88.55          & 87.65           \\
		SECOND-iou* \cite{yan2018second}             & Voxel-based                & 1-stage               & 84.93          & 76.3           & 75.98          & 86.77          & 79.23                  & 77.17          & 87.9                   & 76.3           & 75.95          & 90.23          & 86.61          & 86.37           \\
		TANet \cite{liu2020tanet}                   & Voxel-based                & 1-stage               & 88.17          & 77.75          & 75.31          & -              & -                      & -              & -                      & -              & -              & -              & -              & -               \\
		Part-A2* \cite{shi2020points}                 & Voxel-based                & 2-stage               & 89.55          & 79.41          & 78.85          & \uline{92.15}  & 82.91                  & \uline{82.05}  & 90.2                   & 87.96          & \uline{87.56}  & 92.9           & 90.01          & \uline{88.35}   \\
		Part-A2-free* \cite{shi2020points}           & Voxel-based                & 1-stage               & 89.12          & 78.73          & 77.98          & 91.68          & 80.31                  & 78.1           & 90.1                   & 86.79          & 84.6           & 92.84          & 88.15          & 86.16           \\
		SASSD \cite{he2020structure}                   & Voxel-based                & 1-stage               & \uline{89.69}  & 79.41          & 78.33          & -              & -                      & -              & 90.59                  & 88.43          & 87.49          & -              & -              & -               \\
		Associate-3Det \cite{du2020associate}          & Voxel-based                & 1-stage               & 0              & 79.17          & -              & -              & -                      & -              & -                      & -              & -              & -              & -              & -               \\
		CIASSD \cite{zheng2021cia}                  & Voxel-based                & 1-stage               & 0              & \uline{79.81}  & -              & -              & -                      & -              & -                      & -              & -              & -              & -              & -               \\ 
		\hline
		PV-RCNN* \cite{shi2020pv}                & Point-Voxel                & 2-stage               & 89.26          & 79.16          & \uline{79.39}  & 91.37          & 82.78                  & 80.24          & 89.98                  & 87.7           & 86.59          & 92.72          & 88.59          & 88.04           \\
		Fast Point R-CNN \cite{chen2019fast}        & Point-Voxel                & 2-stage               & 0              & 79             & -              & -              & -                      & -              & -                      & -              & -              & -              & -              & -               \\
		STD \cite{yang2019std}                     & Point-Voxel                & 2-stage               & 0              & 79.8           & -              & -              & -                      & -              & -                      & -              & -              & -              & -              & -               \\
		VIC-Net \cite{jiang2021vic}                 & Point-Voxel                & 1-stage               & 0              & 79.25          & -              & -              & -                      & -              & -                      & -              & -              & -              & -              & -               \\ 
		\hline
		PointRCNN* \cite{shi2019pointrcnn}              & Point-based                & 2-stage               & 88.95          & 78.67          & 77.78          & 91.83          & 80.61                  & 78.18          & 89.92                  & 78.67          & 77.78          & 93.07          & 88.85          & 86.73           \\
		PointRCNN-iou* \cite{shi2019pointrcnn}          & Point-based                & 2-stage               & 89.09          & 78.78          & 78.26          & 91.89          & 80.68                  & 78.41          & 90.19                  & 87.49          & 85.91          & \uline{94.99}  & 88.82          & 86.71           \\
		3DSSD* \cite{yang20203dssd}                  & Point-based                & 1-stage               & 88.82          & 78.58          & 77.47          & -              & -                      & -              & 90.27                  & 87.87          & 86.35          & -              & -              & -               \\
		IA-SSD* \cite{zhang2022not}                 & Point-based                & 1-stage               & 88.78          & 79.12          & 78.12          & 89.52          & 82.86                  & 80.05          & 90.34                  & 88.19          & 86.78          & 93.17          & 89.54          & 88.64           \\
		DBQ-SSD \cite{yang2022dbq}                 & Point-based                & 1-stage               & 0              & 79.56          & -              & -              & -                      & -              & -                      & -              & -              & -              & -              & -               \\
		SPSNet* \cite{liang2023spsnet}                 & Point-based                & 1-stage               & 89.19          & 79.29          & 78.2           & 91.52          & 83.03                  & 80.15          & 90.31                  & 88.72          & 87.31          & 93.2           & 91.21          & 88.9            \\
	\textbf{	PDM-SSD(A)}              & Point-based                & 1-stage               & \textbf{89.12} & \textbf{79.37} & \textbf{78.33} & \textbf{91.57} & \textbf{83.23}         & \textbf{80.37} & \textbf{\uline{92.18}} & \textbf{88.54} & \textbf{86.86} & \textbf{93.17} & \textbf{91.1}  & \textbf{88.67}  \\
		\textbf{PDM-SSD}                 & Point-based                & 1-stage               & \textbf{89.3}  & \textbf{79.75} & \textbf{78.47} & \textbf{91.96} & \textbf{\uline{83.31}} & \textbf{80.59} & \textbf{90.25}         & \textbf{88.64} & \textbf{87.03} & \textbf{93.26} & \textbf{91.24} & \textbf{88.87}  \\
		\bottomrule
	\end{tabular}}
\label{tabel2}
\end{table}

\textbf{Note.} Our model is trained in a single-stage multi-class manner. Thanks to the contributions of mmlab \cite{openpcdet2020}, the PDM-SSD model was trained using their open-source OpenPCDet\footnote{\url{https://github.com/open-mmlab/OpenPCDet}} architecture. Additionally, to show respect and fair competition to the authors of the baseline models, we retested their models with our environment on the \textit{val} set, marked as $X^*$. In the following sections, unless otherwise specified, PDM-SSD refers to the results of joint training, PDM-SSD(J).

\begin{table}
	\centering
	\caption{Quantitative comparison with state-of-the-art methods on the KITTI \textit{val} set for \textit{cyclist} BEV and 3D detection, under the evaluation metric of 3D Average Precision (AP) of 11 and 40 sampling recall points. The best and our PDM-SSD results are highlighted in \textbf{bold} and \uline{underlined}, respectively}
	\resizebox{\textwidth}{!}{
	\begin{tabular}{c|c|c|ccc|ccc|ccc|ccc} 
		\toprule[0.75mm]
		\multirow{3}{*}{Method} & \multirow{3}{*}{Structure} & \multirow{3}{*}{Type} & \multicolumn{6}{c|}{$AP_{3D}@Cyclist$(IoU=0.5)}                                                                   & \multicolumn{6}{c}{$AP_{BEV}@Cyclist$(IoU=0.5)}                                                                            \\ 
		\cline{4-15}
		&                            &                       & \multicolumn{3}{c|}{R11}                                & \multicolumn{3}{c|}{R40}                                 & \multicolumn{3}{c|}{R11}                         & \multicolumn{3}{c}{R40}                                  \\
		&                            &                       & Easy                  & Moderate       & Hard           & Easy                   & Moderate       & Hard           & Easy           & Moderate       & Hard           & Easy           & Moderate               & Hard           \\ 
		\hline
		VoxelNet*\cite{zhou_voxelnet_2017}               & Voxel-based                & 1-stage               & 67.17                 & 47.65          & 45.11          & -                      & -              & -              & 74.41          & 52.18          & 52.49          & -              & -                      & -              \\
		PointPillars*\cite{lang_pointpillars_2019}           & Voxel-based                & 1-stage               & 86.47                 & 68.94          & 66.72          & 89.81                  & 69.71          & 66.91          & 86.66          & 67.21          & 66.72          & 90.01          & 67.45                  & 66.91          \\
		SECOND* \cite{yan2018second}                 & Voxel-based                & 1-stage               & 80.61                 & 67.14          & 63.11          & 82.97                  & 66.74          & 62.78          & 84.02          & 70.7           & 65.48          & 88.04          & 71.16                  & 66.89          \\
		SECOND-iou* \cite{yan2018second}             & Voxel-based                & 1-stage               & 80.44                 & 64.26          & 60.19          & 83.16                  & 63.75          & 60.24          & 84.64          & 66.83          & 63.43          & 86.45          & 67.73                  & 63.38          \\
		TANet \cite{liu2020tanet}                   & Voxel-based                & 1-stage               & 85.98                 & 64.95          & 60.4           & -                      & -              & -              & -              & -              & -              & -              & -                      & -              \\
		Part-A2* \cite{shi2020points}                 & Voxel-based                & 2-stage               & 85.5                  & 69.93          & 65.48          & 90.4                   & 70.1           & 66.9           & 86.88          & 73.32          & 70.84          & 91.93          & 74.6                   & 70.61          \\
		Part-A2-free* \cite{shi2020points}           & Voxel-based                & 2-stage               & 87.95                 & \uline{74.29}  & 69.91          & 91.92                  & 75.33          & 70.59          & 88.75          & \uline{76.31}  & \uline{73.68}  & 93.23          & 78.51                  & \uline{73.94}  \\
		SASSD \cite{he2020structure}                   & Voxel-based                & 1-stage               & 82.8                  & 63.37          & 61.6           & -                      & -              & -              & 86.78          & 71.54          & 65.85          & -              & -                      & -              \\ 
		\hline
		PV-RCNN* \cite{shi2020pv}                & Point-Voxel                & 2-stage               & 84.3                  & 69.29          & 63.59          & 87.05                  & 69.52          & 65.19          & 88.38          & 73.62          & 70.77          & 93.36          & 75.04                  & 70.44          \\ 
		\hline
		PointRCNN* \cite{shi2019pointrcnn}              & Point-based                & 2-stage               & 86.76                 & 71.68          & 65.77          & 91.94                  & 70.98          & 66.72          & 88.41          & 74.31          & 67.96          & 93.91          & 74.71                  & 70.26          \\
		PointRCNN-iou* \cite{shi2019pointrcnn}          & Point-based                & 2-stage               & 86.36                 & 70.99          & 66.23          & 91.48                  & 71.46          & 66.74          & 88.51          & 74.54          & 68.12          & 94.06          & 75.07                  & 70.22          \\
		IA-SSD* \cite{zhang2022not}                 & Point-based                & 1-stage               & 87.28                 & 72.28          & 66.67          & 91.98                  & 72.73          & 68.14          & 88.27          & 73.72          & 70.65          & 93.3           & 74.81                  & 70.68          \\
		SPSNet* \cite{liang2023spsnet}                 & Point-based                & 1-stage               & 88.16                 & 74.01          & \uline{71.2}   & 93.16                  & \uline{75.51}  & \uline{71.1}   & \uline{93.58}  & 75.03          & 72.72          & \uline{95.39}  & 77.57                  & 73.22          \\
		\textbf{PDM-SSD(A)}              & Point-based                & 1-stage               & \textbf{89.12} & \textbf{72.47} & \textbf{67.23} & \textbf{92.43} & \textbf{73.13}         & \textbf{68.47} & \textbf{92.18} & \textbf{73.74} & \textbf{71.16} & \textbf{94.07} & \textbf{81.14}  & \textbf{70.67}  \\
		\textbf{PDM-SSD}                 & Point-based                & 1-stage               & \textbf{\uline{92.04}} & \textbf{72.92} & \textbf{67.88} & \textbf{\uline{93.83}} & \textbf{73.44} & \textbf{68.67} & \textbf{92.99} & \textbf{73.96} & \textbf{71.59} & \textbf{95.24} & \textbf{\uline{85.24}} & \textbf{71.1}  \\
		\bottomrule
	\end{tabular}}
\label{tabel3}
\end{table}

\textbf{Evaluation on KITTI Dataset.} Tables \ref{tabel1} and \ref{tabel2} present the detection performance of PDM-SSD and some state-of-the-art models on \textit{Car} objects in the KITTI \textit{test} and \textit{val} benchmarks. We report their metrics in both 3D and BEV perspectives. In the KITTI benchmark, \textit{Car} objects are divided into three subsets ("Easy," "Moderate," and "Hard") based on difficulty levels. The results on the "Moderate" subset are commonly used as the primary indicator for final ranking. To provide a more intuitive comparison of PDM-SSD's superiority, we categorize the comparative models into three types: point-based, voxel-based, and point-voxel-based, with PDM-SSD belonging to the first category. Table \ref{tabel3} shows the detection results of PDM-SSD on \textit{Cyclist} objects. Table \ref{tabel4} displays the detection metrics of PDM-SSD, IA-SSD, and SPSNet at an IoU threshold of 0.5. Although this threshold is not commonly used for comparing model detection performance, it can reflect the differences in the models' object recall rates.

\begin{table}
	\centering
	\caption{Comparing IA-SSD, SPSNet-SIA and PDM-SSD on the KITTI \textit{val} set for \textit{Car} when IoU threshold is 0.5.}
	\setlength{\extrarowheight}{0pt}
	\addtolength{\extrarowheight}{\aboverulesep}
	\addtolength{\extrarowheight}{\belowrulesep}
	\setlength{\aboverulesep}{0pt}
	\setlength{\belowrulesep}{0pt}
	\resizebox{\textwidth}{!}{
	\begin{tabular}{c|ccc|ccc|ccc|ccc} 
		\toprule
		\multirow{3}{*}{Method}                     & \multicolumn{6}{c}{$AP_{3D}@Car$(IoU=0.5)}         & \multicolumn{6}{c}{$AP_{BEV}@Car$(IoU=0.5)}         \\ 
		\cline{2-13}
		& \multicolumn{3}{c|}{R11} & \multicolumn{3}{c|}{R40} & \multicolumn{3}{c|}{R11} & \multicolumn{3}{c}{R40}   \\
		& Easy  & Moderate & Hard  & Easy  & Moderate & Hard  & Easy  & Moderate & Hard  & Easy  & Moderate & Hard   \\ 
		\hline
		IA-SSD                                      & 90.74 & 90.14    & 89.65 & 96.24 & 95.44    & 94.86 & 90.74 & 90.16    & 89.72 & 96.25 & 95.53    & 95.03  \\
		SPSNet-SIA                                  & 96.19 & 90.17    & 89.69 & 98.04 & 95.5     & 94.99 & 96.24 & 95.81    & 89.75 & 98.07 & 97.28    & 95.11  \\ 
		\hline
		PDM-SSD                                     & 96.46 & 96.11    & 89.64 & 98.3  & 97.39    & 94.86 & 96.42 & 95.96    & 89.74 & 98.3  & 97.38    & 94.86  \\
		\rowcolor[rgb]{0.69,0.69,0.69} - IA-SSD     & +5.72  & +5.97     & -0.01 & +2.06  & +1.95     & 0     & +5.68  & +5.8      & +0.02  & +2.05  & +1.85     & -0.17  \\
		\rowcolor[rgb]{0.69,0.69,0.69} - SPSNet-SIA & +0.27  & +5.94     & -0.05 & +0.26  & +1.89     & -0.13 & +0.18  & +0.15     & -0.01 & +0.23  & +0.1      & -0.25  \\
		\bottomrule
	\end{tabular}}
\label{tabel4}
\end{table}

\textbf{Analysis.} It can be seen that: 
1) In the KITTI \textit{test} split, PDM-SSD performs the best among point-based detectors, with significant improvements of $0.58\% mAP$ and $1.17\% mAP$ over IA-SSD and DBQ-SSD, respectively. We believe that this is a result of PDM-SSD addressing the limited receptive field issue (details in \ref{sec:pdm}). 
2) PDM-SSD has advantages over some voxel-based and point-voxel-based detectors, outperforming Fast Point R-CNN by ($3.45\%, 3.47\%, 5.39\%$) and ($2.2\%, 2.08\%, 4.6\%$) in 3D and bev detection with 40 recall points, respectively. This is exciting because high-speed inference capability of 3D detectors is crucial for autonomous vehicles. 
3) As shown in Table \ref{tabel2}, with only the addition of PDM for auxiliary training, PDM-SSD(A) also achieves considerable gains over IA-SSD, with ($2.05\%, 0.37\%, 0.32\%$) improvement in 3D detection with 40 recall points. This indicates that PDM can help point-wise feature learning for more holistic features of objects, further enhancing the model's learning efficiency. 
4) In addition to the \textit{Car} category, PDM-SSD also provides significant assistance in detecting targets in the \textit{Cyclist} category, as shown in Table \ref{tabel3}. It outperforms IA-SSD by a large margin ($4.76\%, 0.64\%, 1.21\%$) with 11 recall points in terms of $AP_{3D}$, and achieves the best performance in the \textit{easy} level. 
5) As shown in Table \ref{tabel4}, when the IoU is 0.5, PDM-SSD outperforms IA-SSD in all metrics. It improves $AP_{3D}$ by $5.97\%$ and $1.95\%$ at 11 and 40 recall points, respectively, mainly due to the improvement in recall rate. This indicates that the complementarity of heatmap to the voting point set is useful, and recall rate is a prerequisite for a good detector performance. 
6) We conducted a Friedman test analysis \cite{jamaludin2022novel} to evaluate the effectiveness of our work. The final results are as follows: The Friedman test rank was conducted for all point-based detectors for all levels in the KITTI test set with $\alpha = 0.05$ and a degree of freedom of $d_f = 6$. The $p$ for $AP$ is $0.011 (\chi^2 = 16.57)$. As a result, PDM-SSD has an average rank of 1, which is the highest compared to other existing methods for $AP_{3D}$.

We visualized the detection results of three scenes in KITTI \textit{val} sets with PDM-SSD in Fig \ref{fig:detectiveresult}. The figures from left to right are respectively: the image and point clouds with predicted boxes, followed by the space covered by the dilated grid, the ground truth of the heatmap representing the covered space in the scene, and the predicted values of the heatmap. It can be observed that after dilation, the model's learnable space range has significantly increased, achieving almost complete coverage of the target area, especially the center position. The predicted heatmap also successfully identifies all the targets, proving the reliability of our learning method, which plays a positive role in complementing the vote points.

\subsection{Ablation Study}
\label{sec:ablation}

\begin{table}[!h]
	\centering
	\caption{Performance of the model at different SH degrees.}
	\begin{tabular}{c|c|ccc} 
		\toprule
		\multirow{2}{*}{SH degree} & \multirow{2}{*}{numbers} & \multicolumn{3}{c}{R40}                           \\
		&                          & Easy           & Moderate       & Hard            \\ 
		\hline
		2                          & 9                        & 91.1           & 82.89          & 82.29           \\
		3                          & 16                       & \textbf{91.96} & \textbf{83.31} & \textbf{80.59}  \\
		4                          & 25                       & 91.82          & 83.24          & \textbf{80.59}  \\
		\bottomrule
	\end{tabular}
\label{tabel5}
\end{table}

Next, we will conduct ablation studies to evaluate the performance of our proposed modules in PDM-SSD on the KITTI validation split.

\textbf{SH degree.} In the illumination representation model, the higher the degree, the higher the fidelity to real scenes. However, higher degrees also require more coefficients to describe, leading to increased computational complexity. In practical applications, a suitable order is typically chosen to balance accuracy and computational efficiency based on the requirements. Table \ref{tabel5} shows the performance of PDM-SSD with orders 2, 3, and 4. It can be observed that the model performance is comparable between orders 3 and 4, and superior to order 2. Therefore, in order to reduce the number of model parameters and improve computational efficiency, we set the order to 3 with a total of 16 coefficients, which is also a common setting in many rendering models \cite{fridovich2022plenoxels,kerbl20233d}.

\textbf{Coefficients fusion.} In order to maintain the non-linear relationship between the padding feature and the dilation center feature, we adopted a coefficient fusion strategy as shown in Fig. \ref{fig6}. In contrast, directly summing the angle coefficient and the scale coefficient as the feature weight of the new cell, we compared the performance changes brought by these two coefficient fusion strategies in Table. \ref{tabel6}. It can be seen that truncating the features and separately weighting them with the two coefficients can bring higher benefits.

\subsection{Runtime Analysis}
\label{sec:runtime}
\begin{figure}
	\centering
	\includegraphics[width=1\linewidth]{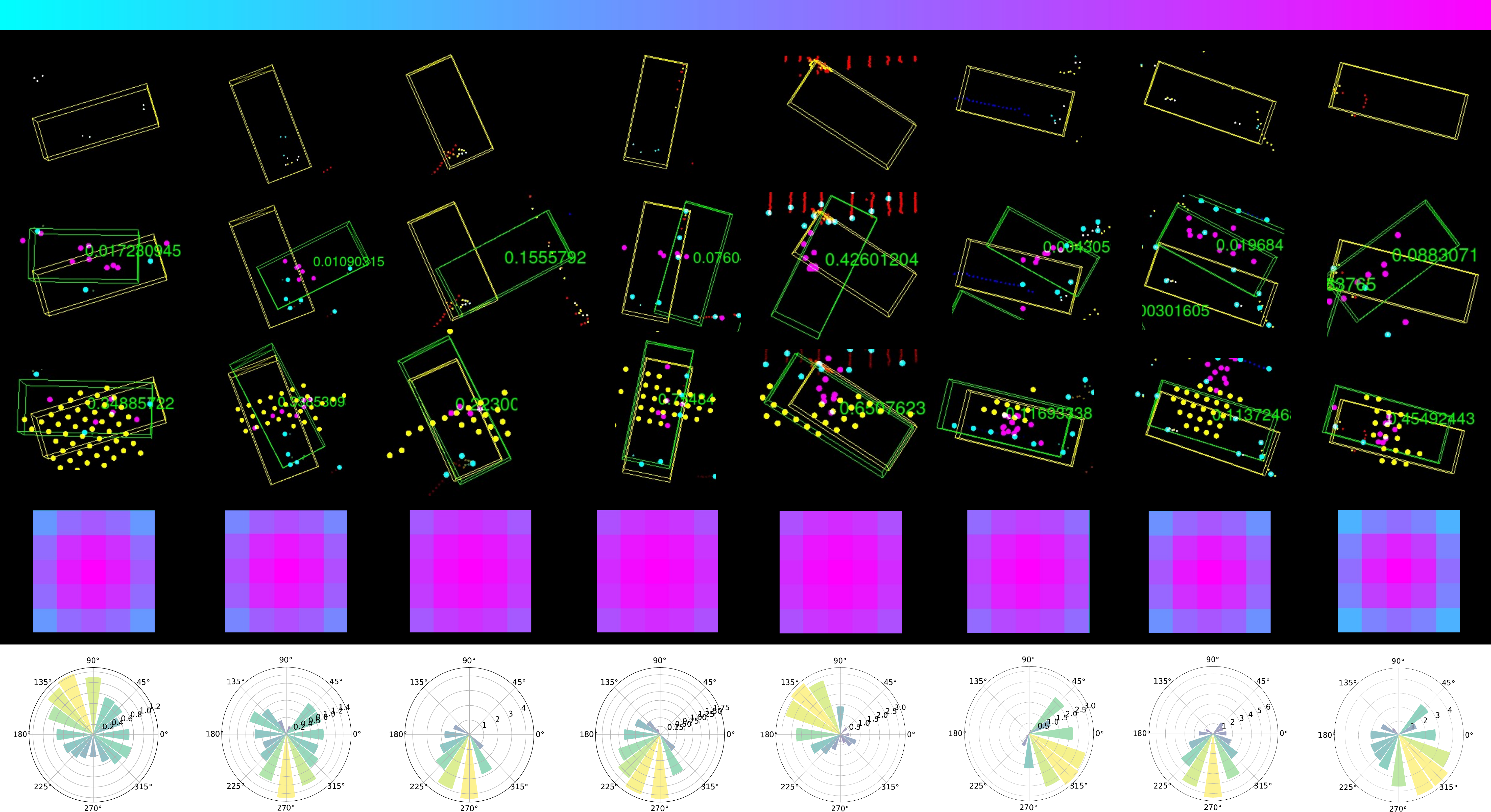}
	\caption{Issue1. The error in the voting point position regressed from the sampling points increases the difficulty of predicting the target box. The first row represents the ground truth bounding box (yellow). The second row shows the inference results of IA-SSD, with green boxes indicating predicted bounding boxes with prediction probabilities, blue points representing selected foreground points, and purple points representing vote points learned from these foreground points. The third row represents the prediction results of our PDM-SSD, with the addition of pseudo foreground points obtained from the predicted heatmap. The fourth row shows the Gaussian coefficients of the foreground points used for final prediction, and the fifth row shows the angle coefficients.}
	\label{fig:issue1}
\end{figure}

\begin{table}[h]
	\centering
	\caption{Model performance under two coefficient fusion methods.}
	\begin{tabular}{c|ccc} 
		\toprule
		\multirow{2}{*}{Fusion mode} & \multicolumn{3}{c}{R40}                           \\
		& Easy           & Moderate       & Hard            \\ 
		\hline
		straight                     & 90.92          & 82.84          & 80.29           \\
		our                          & \textbf{91.96} & \textbf{83.31} & \textbf{80.59}  \\
		\bottomrule
	\end{tabular}
\label{tabel6}
\end{table}

One of the advantages of PDM-SSD is that it uses a point-based 3D backbone, which allows the model to maintain fast inference speed. We analyze the parameter count and GFLOPs of each part of PDM-SSD. PDM-SSD has a very small parameter count, only 3.3M. The main difference between PDM-SSD and many other point-based detectors is the addition of the Neck module, which only contains 0.53M parameters and accounts for only 13\% of the overall GFLOPs, so it has a minimal impact on the model's inference speed. Table. \ref{tabel8} analyzes the inference speed of IA-SSD, PDM-SSD(A), and PDM-SSD. Our experiments were conducted on a single NVIDIA RTX 3090 with Intel i7-12700KF CPU@3.6GHz. The results show that the inference speed of PDM-SSD is 68FPS, slightly lower than the 84FPS of IA-SSD, but it fully meets the hardware limitations of current LiDAR devices and practical application requirements (0.46m/frame at a speed of 120km/h), and it also brings a 0.63\% performance gain. What's even more surprising is that PDM-SSD(A) can maintain the detection speed of 84FPS and bring a significant performance gain. Combining the advantages of PDM in sparse and incomplete object detection undoubtedly improves the safety in practical applications.

\begin{table}
	\centering
	\caption{Analysis of models' inference speed.}
	\begin{tabular}{c|ccc} 
		\toprule
		& Parms (MB) & Speed & Moderate  \\ 
		\hline
		IA-SSD     & 2.7        & 84    & 79.12     \\
		PDM-SSD(A) & 3.3        & 84    & 79.37     \\
		PDM-SSD(J) & 3.3        & 68    & 79.75     \\
		\bottomrule
	\end{tabular}
\label{tabel8}
\end{table}

\subsection{PDM Analysis}
\label{sec:pdm}
\begin{figure}
	\centering
	\includegraphics[width=1\linewidth]{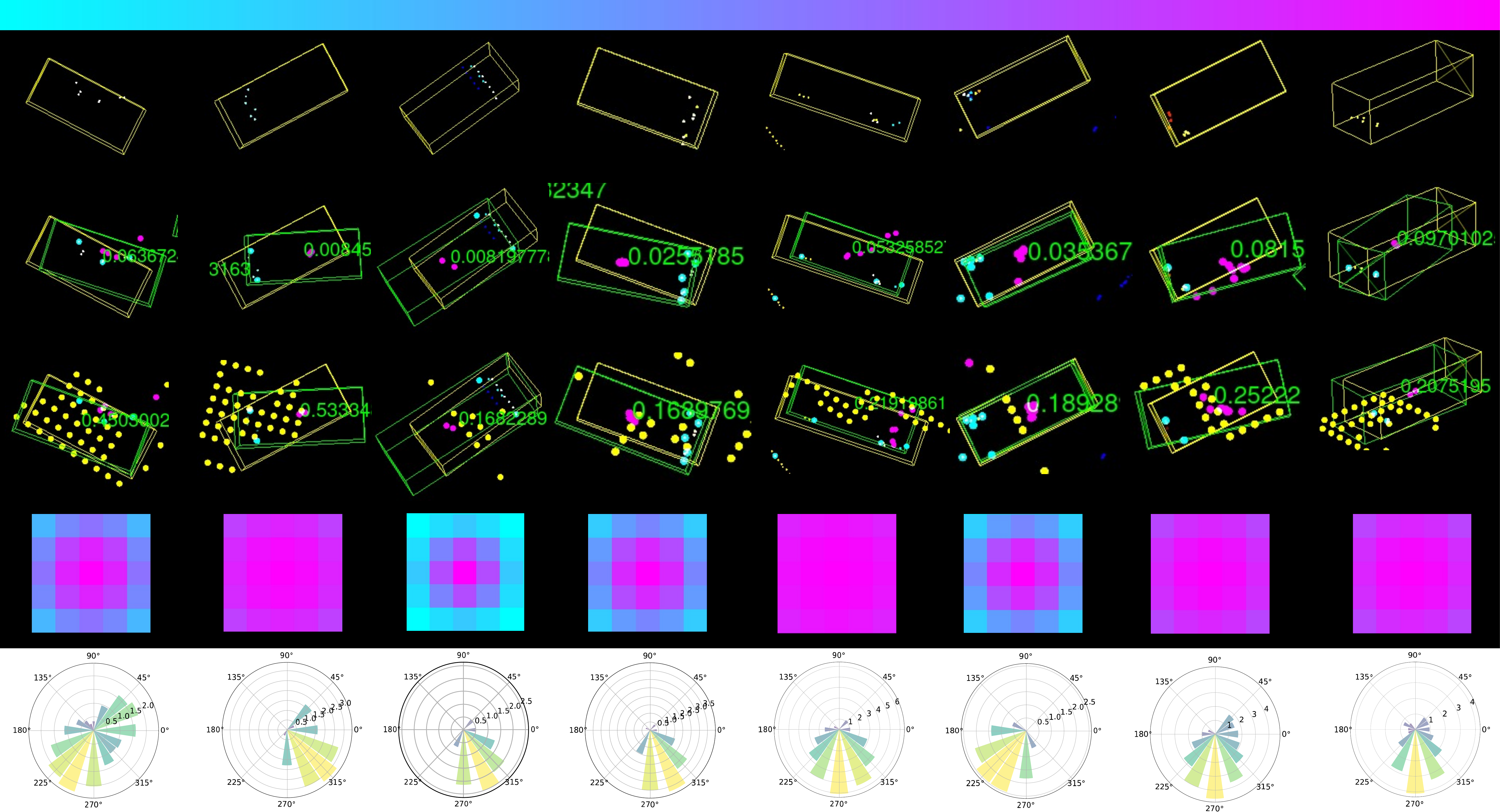}
	\caption{Issue2. The detection box regressed from the features learned from context has a probability lower than the threshold. The first row represents the ground truth bounding box (yellow). The second row shows the inference results of IA-SSD, with green boxes indicating predicted bounding boxes with prediction probabilities, blue points representing selected foreground points, and purple points representing vote points learned from these foreground points. The third row represents the prediction results of our PDM-SSD, with the addition of pseudo foreground points obtained from the predicted heatmap. The fourth row shows the Gaussian coefficients of the foreground points used for final prediction, and the fifth row shows the angle coefficients.}
	\label{fig:issue2}
\end{figure}

In the previous sections, we have quantitatively analyzed the advantages of PDM-SSD in terms of overall performance (\ref{sec:comparison}) and runtime (\ref{sec:runtime}). In this section, we will analyze the role of PDM at the object level, particularly in detecting sparse and incomplete targets. As mentioned earlier, current point-based detectors suffer from two important issues due to discontinuous receptive fields. \textbf{\uppercase\expandafter{\romannumeral1}:} The error in the voting point position regressed from the sampling points increases the difficulty of predicting the target box. \textbf{\uppercase\expandafter{\romannumeral2}:} The detection box regressed from the features learned from context has a probability lower than the threshold. 
We will analyze the contributions made by PDM-SSD to address these two issues separately. 

\textbf{Issue \uppercase\expandafter{\romannumeral1}.} The features predicted by querying the vote points are crucial for determining the class and geometric parameters of the target bounding box. The position of the vote points has a significant impact on the prediction results. For sparse and particularly incomplete targets, the model's receptive field is limited when learning from only occupied points. It is difficult for the model to actively learn the overall features of the target and establish connections between points. This can easily lead to the deviation of vote points from the target center, resulting in a low IoU of the predicted bounding box compared to the threshold. In contrast, PDM-SSD learns from the dilated grid features to GT heatmap, which promotes the learning of global features by the dilated centers. Additionally, the predicted heatmap can complement the vote points, which we believe can alleviate the issue of large positional errors in vote points.

We illustrate this situation with some examples in Figure \ref{fig:issue1}. In each column of the figure, there is a sparse or incomplete target. The first row represents the ground truth bounding box (yellow). The second row shows the inference results of IA-SSD, with green boxes indicating predicted bounding boxes with prediction probabilities, blue points representing selected foreground points, and purple points representing vote points learned from these foreground points. The third row represents the prediction results of our PDM-SSD, with the addition of pseudo foreground points obtained from the predicted heatmap. The fourth row shows the Gaussian coefficients of the foreground points used for final prediction, and the fifth row shows the angle coefficients. From the figure, it can be observed that the vote points regressed by IA-SSD deviate from the center of the ground truth bounding box, while PDM-SSD regresses the vote points more accurately, leading to more accurate predictions. It is worth noting that in the seventh column, both IA-SSD and PDM-SSD regress similar vote point positions, which deviate from the ideal position. However, PDM-SSD still achieves good regression results with the addition of pseudo foreground points, while IA-SSD, due to only being able to utilize existing points, cannot make more accurate selections to significantly deviate the predicted bounding box from the ground truth. These examples provide a visual demonstration of the improvements made by PDM-SSD in addressing \textbf{Issue \uppercase\expandafter{\romannumeral1}}.

\textbf{Issue \uppercase\expandafter{\romannumeral2}.} The prediction results of the detector are determined by both IoU and prediction probability. This means that even if the vote points have small positioning errors and the predicted box is close to the ground truth in terms of position and size, the prediction result may still be excluded by the detector if the target probability is low. This is a common problem in the original point-based detector because object detection is essentially a combination of classification and regression tasks. In regression tasks, the spatial geometry information of points is more important, and with a large amount of training data, the model can easily learn accurate detection boxes from discrete and sparse points. However, for classification tasks, the semantic information of points is more important, and the limited receptive field at this time cannot make the point-wise features contain the overall information of the target, thus unable to obtain accurate semantic information of the target box. In the learning process, PDM-SSD connects the inflated centers through coefficients fusion and height compression, and obtains more global information from grid features through joint learning. We believe that it can improve the prediction probability for this type of target.

In Figure \ref{fig:issue2}, we demonstrate this situation where the IA-SSD accurately predicts the positions of vote points regressed from foreground points in the second row, but the target probabilities are all below the threshold (0.1). This undoubtedly affects the model's recall rate. However, PDM-SSD can achieve higher prediction probabilities, reducing the impact of this issue on the overall performance of the model.

\section{Dicussion}
\label{sec:dicussion}
PDM-SSD provides a new approach to address the issue of discontinuous receptive fields in point-based detectors. It eliminates the complex process of integrating point and grid representations in the backbone and instead utilizes PDM to directly lift and fill features for sampled points within the neck module. This not only maintains a lightweight model but also exploits the advantages of both representations.
We have extensively demonstrated the superiority of PDM-SSD through numerous experiments, as described in Section \ref{sec:experiments}. In terms of detection accuracy, PDM-SSD surpasses the state-of-the-art point-based detectors and can compete with some voxel-based models. In terms of inference speed, PDM-SSD can run efficiently at 68 FPS, which is much higher than the scanning frequency of current LiDARs, making it suitable for practical applications. The auxiliary trained PDM-SSD (A) achieves even higher detection accuracy without sacrificing inference efficiency. Additionally, the model parameters of PDM-SSD are only 3.3MB, greatly reducing the deployment difficulty. In the object-level experimental analysis, we found that PDM-SSD effectively mitigates the issues of large vote point errors and low object box prediction probabilities caused by limited receptive fields in current point-based detectors. Overall, PDM-SSD has indeed identified new problems and made further advancements in the current research, demonstrating its value.

\textbf{Limitations and outlook.} Objectively speaking, PDM-SSD still has the following issues: 1) We avoided the detection of pedestrians in the KITTI dataset because the small volume and limited impact of pedestrians make the $5\times 5$ structural element unsuitable for detecting such targets. In future work, we will select different sizes of $B$ for different classes of objects. 2) In the scale coefficient of feature padding, we set two variables as independently and identically distributed, following the setting of heatmap in CenterNet \cite{yin2021center}. However, we believe this is not optimal, and in future work, we will split the Gaussian into a mixture of scale and rotation to learn the covariance and embed more geometric information into the learning process. 3) The final context learning module we used is still PointNet, which may not fully utilize the mixed features provided by the mixed head. In future work, we will design more sophisticated learning modules.

\section{Conclusion}
\label{sec:conclusion}
In this article, we propose single stage point-based 3D object detector called PDM-SSD. Our goal is to alleviate the limited receptive field issue while maintaining the fast inference speed of point-based detectors. Currently, point-based detectors can only learn features from existing points, and their receptive field is discontinuous and limited when the query radius expands, especially for sparse or extremely incomplete objects. This not only leads to large position errors in regression vote points but also results in prediction probabilities for object boxes lower than the threshold. PDM-SSD expands the learning space of the model through Point Dilation, specifically covering the unoccupied space near the center of the object in the original point cloud. Then, it fills these spaces with features that can be backpropagated, and the information from multiple dilation centers is connected through height compression. Finally, these features are jointly learned through a hybrid head. The experimental results show that PDM-SSD achieves competitive performance in terms of detection accuracy, surpassing all current point-based models in multiple metrics. In terms of inference speed, it fully meets the current application requirements. More importantly, from the object-level experiments, we can intuitively see the contributions of PDM-SSD in addressing the issues of current point-based detectors.

\section*{Acknowledgments}
This document is the results of the research project funded by the CAS Innovation Fund (E01Z040101)
\bibliographystyle{unsrt}  
\bibliography{references}

\begin{thebibliography}{10}

\bibitem{geiger2012we}
Andreas Geiger, Philip Lenz, and Raquel Urtasun.
\newblock Are we ready for autonomous driving? the kitti vision benchmark suite.
\newblock In {\em 2012 IEEE conference on computer vision and pattern recognition}, pages 3354--3361. IEEE, 2012.

\bibitem{sun2020scalability}
Pei Sun, Henrik Kretzschmar, Xerxes Dotiwalla, Aurelien Chouard, Vijaysai Patnaik, Paul Tsui, James Guo, Yin Zhou, Yuning Chai, Benjamin Caine, et~al.
\newblock Scalability in perception for autonomous driving: Waymo open dataset.
\newblock In {\em Proceedings of the IEEE/CVF conference on computer vision and pattern recognition}, pages 2446--2454, 2020.

\bibitem{fan2021rangedet}
Lue Fan, Xuan Xiong, Feng Wang, Naiyan Wang, and Zhaoxiang Zhang.
\newblock Rangedet: In defense of range view for lidar-based 3d object detection.
\newblock In {\em Proceedings of the IEEE/CVF International Conference on Computer Vision}, pages 2918--2927, 2021.

\bibitem{simon_complex-yolo_2018}
Martin Simon, Stefan Milz, Karl Amende, and Horst-Michael Gross.
\newblock Complex-{YOLO}: Real-time 3d object detection on point clouds.

\bibitem{noauthor_multi-view_nodate}
Multi-view 3d object detection network for autonomous driving.

\bibitem{beltran2018birdnet}
Jorge Beltr{\'a}n, Carlos Guindel, Francisco~Miguel Moreno, Daniel Cruzado, Fernando Garcia, and Arturo De~La~Escalera.
\newblock Birdnet: a 3d object detection framework from lidar information.
\newblock In {\em 2018 21st International Conference on Intelligent Transportation Systems (ITSC)}, pages 3517--3523. IEEE, 2018.

\bibitem{zeng2018rt3d}
Yiming Zeng, Yu~Hu, Shice Liu, Jing Ye, Yinhe Han, Xiaowei Li, and Ninghui Sun.
\newblock Rt3d: Real-time 3-d vehicle detection in lidar point cloud for autonomous driving.
\newblock {\em IEEE Robotics and Automation Letters}, 3(4):3434--3440, 2018.

\bibitem{ali2018yolo3d}
Waleed Ali, Sherif Abdelkarim, Mahmoud Zidan, Mohamed Zahran, and Ahmad El~Sallab.
\newblock Yolo3d: End-to-end real-time 3d oriented object bounding box detection from lidar point cloud.
\newblock In {\em Proceedings of the European conference on computer vision (ECCV) workshops}, pages 0--0, 2018.

\bibitem{barrera2020birdnet+}
Alejandro Barrera, Carlos Guindel, Jorge Beltr{\'a}n, and Fernando Garc{\'\i}a.
\newblock Birdnet+: End-to-end 3d object detection in lidar bird’s eye view.
\newblock In {\em 2020 IEEE 23rd International Conference on Intelligent Transportation Systems (ITSC)}, pages 1--6. IEEE, 2020.

\bibitem{CHEN2023110952}
Mu~Chen, Pengfei Liu, and Huaici Zhao.
\newblock Bcaf-3d: Bilateral content awareness fusion for cross-modal 3d object detection.
\newblock {\em Knowledge-Based Systems}, 279:110952, 2023.

\bibitem{zheng2021se}
Wu~Zheng, Weiliang Tang, Li~Jiang, and Chi-Wing Fu.
\newblock Se-ssd: Self-ensembling single-stage object detector from point cloud.
\newblock In {\em Proceedings of the IEEE/CVF Conference on Computer Vision and Pattern Recognition}, pages 14494--14503, 2021.

\bibitem{zhang2022not}
Yifan Zhang, Qingyong Hu, Guoquan Xu, Yanxin Ma, Jianwei Wan, and Yulan Guo.
\newblock Not all points are equal: Learning highly efficient point-based detectors for 3d lidar point clouds.
\newblock In {\em Proceedings of the IEEE/CVF Conference on Computer Vision and Pattern Recognition}, pages 18953--18962, 2022.

\bibitem{shi2019pointrcnn}
Shaoshuai Shi, Xiaogang Wang, and Hongsheng Li.
\newblock Pointrcnn: 3{D} object proposal generation and detection from point cloud.
\newblock In {\em CVPR}, pages 770--779, 2019.

\bibitem{shi2020point}
Weijing Shi and Raj Rajkumar.
\newblock Point-gnn: Graph neural network for 3{D} object detection in a point cloud.
\newblock In {\em CVPR}, pages 1711--1719, 2020.

\bibitem{yang20203dssd}
Zetong Yang, Yanan Sun, Shu Liu, and Jiaya Jia.
\newblock 3dssd: Point-based 3{D} single stage object detector.
\newblock In {\em CVPR}, pages 11040--11048, 2020.

\bibitem{hu2021learning}
Qingyong Hu, Bo~Yang, Linhai Xie, Stefano Rosa, Yulan Guo, Zhihua Wang, Niki Trigoni, and Andrew Markham.
\newblock Learning semantic segmentation of large-scale point clouds with random sampling.
\newblock {\em IEEE Transactions on Pattern Analysis and Machine Intelligence}, 2021.

\bibitem{hu2021sqn}
Qingyong Hu, Bo~Yang, Guangchi Fang, Yulan Guo, Ales Leonardis, Niki Trigoni, and Andrew Markham.
\newblock Sqn: Weakly-supervised semantic segmentation of large-scale {3d} point clouds with 1000x fewer labels.
\newblock {\em arXiv preprint arXiv:2104.04891}, 2021.

\bibitem{hu2022sensaturban}
Qingyong Hu, Bo~Yang, Sheikh Khalid, Wen Xiao, Niki Trigoni, and Andrew Markham.
\newblock Sensaturban: Learning semantics from urban-scale photogrammetric point clouds.
\newblock {\em International Journal of Computer Vision}, pages 1--28, 2022.

\bibitem{wei2022spatial}
Yimin Wei, Hao Liu, Tingting Xie, Qiuhong Ke, and Yulan Guo.
\newblock Spatial-temporal transformer for {3d} point cloud sequences.
\newblock In {\em WACV}, pages 1171--1180, 2022.

\bibitem{yin2021center}
Tianwei Yin, Xingyi Zhou, and Philipp Krahenbuhl.
\newblock Center-based 3d object detection and tracking.
\newblock In {\em Proceedings of the IEEE/CVF conference on computer vision and pattern recognition}, pages 11784--11793, 2021.

\bibitem{zhou_voxelnet_2017}
Yin Zhou and Oncel Tuzel.
\newblock {VoxelNet}: End-to-end learning for point cloud based 3d object detection.

\bibitem{chen2023voxelnext}
Yukang Chen, Jianhui Liu, Xiangyu Zhang, Xiaojuan Qi, and Jiaya Jia.
\newblock Voxelnext: Fully sparse voxelnet for 3d object detection and tracking.
\newblock In {\em Proceedings of the IEEE/CVF Conference on Computer Vision and Pattern Recognition}, pages 21674--21683, 2023.

\bibitem{wang2023distillbev}
Zeyu Wang, Dingwen Li, Chenxu Luo, Cihang Xie, and Xiaodong Yang.
\newblock Distillbev: Boosting multi-camera 3d object detection with cross-modal knowledge distillation.
\newblock In {\em Proceedings of the IEEE/CVF International Conference on Computer Vision}, pages 8637--8646, 2023.

\bibitem{ye_lidarmultinet_2022}
Dongqiangzi Ye, Zixiang Zhou, Weijia Chen, Yufei Xie, Yu~Wang, Panqu Wang, and Hassan Foroosh.
\newblock {LidarMultiNet}: Towards a unified multi-task network for {LiDAR} perception.

\bibitem{lang_pointpillars_2019}
Alex~H. Lang, Sourabh Vora, Holger Caesar, Lubing Zhou, Jiong Yang, and Oscar Beijbom.
\newblock {PointPillars}: Fast encoders for object detection from point clouds.

\bibitem{shi_pillarnet_2022}
Guangsheng Shi, Ruifeng Li, and Chao Ma.
\newblock {PillarNet}: Real-time and high-performance pillar-based 3d object detection.

\bibitem{zhou_fastpillars_2023}
Sifan Zhou, Zhi Tian, Xiangxiang Chu, Xinyu Zhang, Bo~Zhang, Xiaobo Lu, Chengjian Feng, Zequn Jie, Patrick~Yin Chiang, and Lin Ma.
\newblock {FastPillars}: A deployment-friendly pillar-based 3d detector.

\bibitem{jia2023driveadapter}
Xiaosong Jia, Yulu Gao, Li~Chen, Junchi Yan, Patrick~Langechuan Liu, and Hongyang Li.
\newblock Driveadapter: Breaking the coupling barrier of perception and planning in end-to-end autonomous driving.
\newblock In {\em Proceedings of the IEEE/CVF International Conference on Computer Vision}, pages 7953--7963, 2023.

\bibitem{hu2023planning}
Yihan Hu, Jiazhi Yang, Li~Chen, Keyu Li, Chonghao Sima, Xizhou Zhu, Siqi Chai, Senyao Du, Tianwei Lin, Wenhai Wang, et~al.
\newblock Planning-oriented autonomous driving.
\newblock In {\em Proceedings of the IEEE/CVF Conference on Computer Vision and Pattern Recognition}, pages 17853--17862, 2023.

\bibitem{li2023delving}
Hongyang Li, Chonghao Sima, Jifeng Dai, Wenhai Wang, Lewei Lu, Huijie Wang, Jia Zeng, Zhiqi Li, Jiazhi Yang, Hanming Deng, et~al.
\newblock Delving into the devils of bird's-eye-view perception: A review, evaluation and recipe.
\newblock {\em IEEE Transactions on Pattern Analysis and Machine Intelligence}, 2023.

\bibitem{li2022bevformer}
Zhiqi Li, Wenhai Wang, Hongyang Li, Enze Xie, Chonghao Sima, Tong Lu, Yu~Qiao, and Jifeng Dai.
\newblock Bevformer: Learning bird’s-eye-view representation from multi-camera images via spatiotemporal transformers.
\newblock In {\em European conference on computer vision}, pages 1--18. Springer, 2022.

\bibitem{meyer2019lasernet}
Gregory~P Meyer, Ankit Laddha, Eric Kee, Carlos Vallespi-Gonzalez, and Carl~K Wellington.
\newblock Lasernet: An efficient probabilistic 3d object detector for autonomous driving.
\newblock In {\em Proceedings of the IEEE/CVF conference on computer vision and pattern recognition}, pages 12677--12686, 2019.

\bibitem{yu2018deep}
Fisher Yu, Dequan Wang, Evan Shelhamer, and Trevor Darrell.
\newblock Deep layer aggregation.
\newblock In {\em Proceedings of the IEEE conference on computer vision and pattern recognition}, pages 2403--2412, 2018.

\bibitem{meyer2020laserflow}
Gregory~P Meyer, Jake Charland, Shreyash Pandey, Ankit Laddha, Shivam Gautam, Carlos Vallespi-Gonzalez, and Carl~K Wellington.
\newblock Laserflow: Efficient and probabilistic object detection and motion forecasting.
\newblock {\em IEEE Robotics and Automation Letters}, 6(2):526--533, 2020.

\bibitem{liang2020rangercnn}
Zhidong Liang, Ming Zhang, Zehan Zhang, Xian Zhao, and Shiliang Pu.
\newblock Rangercnn: Towards fast and accurate 3d object detection with range image representation.
\newblock {\em arXiv preprint arXiv:2009.00206}, 2020.

\bibitem{qi_deep_2019}
Charles~R. Qi, Or~Litany, Kaiming He, and Leonidas~J. Guibas.
\newblock Deep hough voting for 3d object detection in point clouds.

\bibitem{yang_3dssd_2020}
Zetong Yang, Yanan Sun, Shu Liu, and Jiaya Jia.
\newblock 3dssd: Point-based 3d single stage object detector.

\bibitem{qi2017pointnet}
Charles~R Qi, Hao Su, Kaichun Mo, and Leonidas~J Guibas.
\newblock Pointnet: Deep learning on point sets for 3d classification and segmentation.
\newblock In {\em Proceedings of the IEEE conference on computer vision and pattern recognition}, pages 652--660, 2017.

\bibitem{yang2022dbq}
Jinrong Yang, Lin Song, Songtao Liu, Weixin Mao, Zeming Li, Xiaoping Li, Hongbin Sun, Jian Sun, and Nanning Zheng.
\newblock Dbq-ssd: Dynamic ball query for efficient 3d object detection.
\newblock In {\em The Eleventh International Conference on Learning Representations}, 2022.

\bibitem{liang2023spsnet}
Ao~Liang, Hao Zhang, Haiyang Hua, Whenyu Chen, and Huaici Zhao.
\newblock Spsnet: Boosting 3d point-based object detectors with stable point sampling.
\newblock {\em Engineering Applications of Artificial Intelligence}, 126:106807, 2023.

\bibitem{zhu2023understanding}
Zijian Zhu, Yichi Zhang, Hai Chen, Yinpeng Dong, Shu Zhao, Wenbo Ding, Jiachen Zhong, and Shibao Zheng.
\newblock Understanding the robustness of 3d object detection with bird's-eye-view representations in autonomous driving.
\newblock In {\em Proceedings of the IEEE/CVF Conference on Computer Vision and Pattern Recognition}, pages 21600--21610, 2023.

\bibitem{liu2019point}
Zhijian Liu, Haotian Tang, Yujun Lin, and Song Han.
\newblock Point-voxel cnn for efficient 3d deep learning.
\newblock {\em Advances in Neural Information Processing Systems}, 32, 2019.

\bibitem{tang2020searching}
Haotian Tang, Zhijian Liu, Shengyu Zhao, Yujun Lin, Ji~Lin, Hanrui Wang, and Song Han.
\newblock Searching efficient 3d architectures with sparse point-voxel convolution.
\newblock In {\em Computer Vision--ECCV 2020: 16th European Conference, Glasgow, UK, August 23--28, 2020, Proceedings, Part XXVIII}, pages 685--702. Springer, 2020.

\bibitem{shi2023pv}
Shaoshuai Shi, Li~Jiang, Jiajun Deng, Zhe Wang, Chaoxu Guo, Jianping Shi, Xiaogang Wang, and Hongsheng Li.
\newblock Pv-rcnn++: Point-voxel feature set abstraction with local vector representation for 3d object detection.
\newblock {\em International Journal of Computer Vision}, 131(2):531--551, 2023.

\bibitem{yan2018second}
Yan Yan, Yuxing Mao, and Bo~Li.
\newblock Second: Sparsely embedded convolutional detection.
\newblock {\em Sensors}, 18(10):3337, 2018.

\bibitem{qian2022badet}
Rui Qian, Xin Lai, and Xirong Li.
\newblock Badet: Boundary-aware 3d object detection from point clouds.
\newblock {\em Pattern Recognition}, 125:108524, 2022.

\bibitem{wu2010morphological}
Jiaji Wu, Anand Paul, Yan Xing, Yong Fang, Jechang Jeong, Licheng Jiao, and Guangming Shi.
\newblock Morphological dilation image coding with context weights prediction.
\newblock {\em Signal Processing: Image Communication}, 25(10):717--728, 2010.

\bibitem{zhou2019iou}
Dingfu Zhou, Jin Fang, Xibin Song, Chenye Guan, Junbo Yin, Yuchao Dai, and Ruigang Yang.
\newblock Iou loss for 2d/3d object detection.
\newblock In {\em 2019 international conference on 3D vision (3DV)}, pages 85--94. IEEE, 2019.

\bibitem{liu2020tanet}
Zhe Liu, Xin Zhao, Tengteng Huang, Ruolan Hu, Yu~Zhou, and Xiang Bai.
\newblock Tanet: Robust 3d object detection from point clouds with triple attention.
\newblock In {\em Proceedings of the AAAI conference on artificial intelligence}, volume~34, pages 11677--11684, 2020.

\bibitem{shi2020points}
Shaoshuai Shi, Zhe Wang, Jianping Shi, Xiaogang Wang, and Hongsheng Li.
\newblock From points to parts: 3d object detection from point cloud with part-aware and part-aggregation network.
\newblock {\em IEEE transactions on pattern analysis and machine intelligence}, 43(8):2647--2664, 2020.

\bibitem{zheng2021cia}
Wu~Zheng, Weiliang Tang, Sijin Chen, Li~Jiang, and Chi-Wing Fu.
\newblock Cia-ssd: Confident iou-aware single-stage object detector from point cloud.
\newblock In {\em Proceedings of the AAAI conference on artificial intelligence}, volume~35, pages 3555--3562, 2021.

\bibitem{he2020structure}
Chenhang He, Hui Zeng, Jianqiang Huang, Xian-Sheng Hua, and Lei Zhang.
\newblock Structure aware single-stage 3d object detection from point cloud.
\newblock In {\em Proceedings of the IEEE/CVF conference on computer vision and pattern recognition}, pages 11873--11882, 2020.

\bibitem{du2020associate}
Liang Du, Xiaoqing Ye, Xiao Tan, Jianfeng Feng, Zhenbo Xu, Errui Ding, and Shilei Wen.
\newblock Associate-3ddet: Perceptual-to-conceptual association for 3d point cloud object detection.
\newblock In {\em Proceedings of the IEEE/CVF conference on computer vision and pattern recognition}, pages 13329--13338, 2020.

\bibitem{he2022svga}
Qingdong He, Zhengning Wang, Hao Zeng, Yi~Zeng, and Yijun Liu.
\newblock Svga-net: Sparse voxel-graph attention network for 3d object detection from point clouds.
\newblock In {\em Proceedings of the AAAI Conference on Artificial Intelligence}, volume~36, pages 870--878, 2022.

\bibitem{chen2019fast}
Yilun Chen, Shu Liu, Xiaoyong Shen, and Jiaya Jia.
\newblock Fast point r-cnn.
\newblock In {\em Proceedings of the IEEE/CVF international conference on computer vision}, pages 9775--9784, 2019.

\bibitem{yang2019std}
Zetong Yang, Yanan Sun, Shu Liu, Xiaoyong Shen, and Jiaya Jia.
\newblock Std: Sparse-to-dense 3d object detector for point cloud.
\newblock In {\em Proceedings of the IEEE/CVF international conference on computer vision}, pages 1951--1960, 2019.

\bibitem{shi2020pv}
Shaoshuai Shi, Chaoxu Guo, Li~Jiang, Zhe Wang, Jianping Shi, Xiaogang Wang, and Hongsheng Li.
\newblock Pv-rcnn: Point-voxel feature set abstraction for 3d object detection.
\newblock In {\em Proceedings of the IEEE/CVF Conference on Computer Vision and Pattern Recognition}, pages 10529--10538, 2020.

\bibitem{yang2022unified}
Zetong Yang, Li~Jiang, Yanan Sun, Bernt Schiele, and Jiaya Jia.
\newblock A unified query-based paradigm for point cloud understanding.
\newblock In {\em Proceedings of the IEEE/CVF Conference on Computer Vision and Pattern Recognition}, pages 8541--8551, 2022.

\bibitem{jiang2021vic}
Tianyuan Jiang, Nan Song, Huanyu Liu, Ruihao Yin, Ye~Gong, and Jian Yao.
\newblock Vic-net: Voxelization information compensation network for point cloud 3d object detection.
\newblock In {\em 2021 IEEE International Conference on Robotics and Automation (ICRA)}, pages 13408--13414. IEEE, 2021.

\bibitem{noh2021hvpr}
Jongyoun Noh, Sanghoon Lee, and Bumsub Ham.
\newblock Hvpr: Hybrid voxel-point representation for single-stage 3d object detection.
\newblock In {\em Proceedings of the IEEE/CVF conference on computer vision and pattern recognition}, pages 14605--14614, 2021.

\bibitem{li20203d}
Jiale Li, Shujie Luo, Ziqi Zhu, Hang Dai, Andrey~S Krylov, Yong Ding, and Ling Shao.
\newblock 3d iou-net: Iou guided 3d object detector for point clouds.
\newblock {\em arXiv preprint arXiv:2004.04962}, 2020.

\bibitem{openpcdet2020}
OpenPCDet~Development Team.
\newblock Openpcdet: An open-source toolbox for 3d object detection from point clouds.
\newblock \url{https://github.com/open-mmlab/OpenPCDet}, 2020.

\bibitem{jamaludin2022novel}
Siti Zulaikha~Mohd Jamaludin, Nurul~Atiqah Romli, Mohd Shareduwan~Mohd Kasihmuddin, Aslina Baharum, Mohd~Asyraf Mansor, and Muhammad~Fadhil Marsani.
\newblock Novel logic mining incorporating log linear approach.
\newblock {\em Journal of King Saud University-Computer and Information Sciences}, 34(10):9011--9027, 2022.

\bibitem{fridovich2022plenoxels}
Sara Fridovich-Keil, Alex Yu, Matthew Tancik, Qinhong Chen, Benjamin Recht, and Angjoo Kanazawa.
\newblock Plenoxels: Radiance fields without neural networks.
\newblock In {\em Proceedings of the IEEE/CVF Conference on Computer Vision and Pattern Recognition}, pages 5501--5510, 2022.

\bibitem{kerbl20233d}
Bernhard Kerbl, Georgios Kopanas, Thomas Leimk{\"u}hler, and George Drettakis.
\newblock 3d gaussian splatting for real-time radiance field rendering.
\newblock {\em ACM Transactions on Graphics (ToG)}, 42(4):1--14, 2023.

\end{thebibliography}

\end{document}